\pdfoutput=1
\documentclass[11pt]{article}

\usepackage[preprint]{acl}

\usepackage{times}
\usepackage{latexsym}

\usepackage[T1]{fontenc}
\usepackage[utf8]{inputenc}

\usepackage{microtype}

\usepackage{inconsolata}
\usepackage{multirow}
\usepackage{microtype}
\usepackage{graphicx}
\usepackage{tabularx}
\usepackage{subfigure}
\usepackage{booktabs} % for professional tables
\usepackage{amsmath}
\usepackage{xspace}
\newcommand{\ourset}{{Mementos}\xspace}

\usepackage{soul}

\usepackage[export]{adjustbox}
\newcommand{\MyEmoji}[1]{\includegraphics[width=1em,valign=t]{#1}}
\newcommand{\fimeframe}{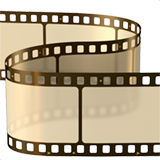}

\title{\MyEmoji{\fimeframe} Mementos: A Comprehensive Benchmark for Multimodal Large Language Model Reasoning over Image Sequences}

\author{Xiyao Wang$^{1 \dag}$\quad  Yuhang Zhou$^{1}$\quad   Xiaoyu Liu$^{1}$\\ 
\textbf{Hongjin Lu$^{1}$}\quad   \textbf{Yuancheng Xu$^{1}$}\quad   \textbf{Feihong He$^{2}$}\quad  \textbf{Jaehong Yoon$^{2}$}\quad \textbf{Taixi Lu$^{2}$}\\ 
\textbf{Gedas Bertasius$^{2}$}\quad  \textbf{Mohit Bansal$^{2}$}\quad   \textbf{Huaxiu Yao$^{2 \ddag}$}\quad   \textbf{Furong Huang$^{1 \ddag}$} \\
  $^1$University of Maryland, College Park \\
  $^2$UNC-Chapel Hill, Chapel Hill \\
  $^\dag$\texttt{xywang@umd.edu} $^\ddag$ Equal advising}
  
\begin{document}
\maketitle
\begin{abstract}
% Large Multimodal Models (MLLMs) have recently shown strong multi-modal task handling capabilities. 
% However, MLLMs still struggle with accurate visual reasoning. 
% Existing benchmarks and methods mainly focus on assessing and enhancing MLLM's ability on single image reasoning, which concentrates on knowledge reasoning of static objects, while reasoning capability over image sequences, which requires MLLM to analyze object position across images and infer the dynamic behavior, is more challenging and crucial in real-world applications. 
% In this paper, we introduce a novel and challenging benchmark named \ourset for evaluating MLLM's reasoning abilities with sequential image inputs. 
% \ourset consists of 4,711 image sequences from diverse scenarios with varying episode lengths. 
% We also propose a GPT-4V assisted evaluation procedure to aid in assessing MLLMs' reasoning performance. 
% After evaluating 7 most recent MLLMs on \ourset, including GPT-4V and Gemini, we find that existing MLLMs lack strong reasoning capabilities when facing with image sequence inputs. 
% Without external information prompts, MLLMs struggle to accurately describe episodes in given image sequences, leading to significant object and behavioral hallucinations in their generated descriptions. 
% Through quantitative analysis and case studies, we identify three factors affecting the reasoning ability of existing MLLMs for image sequences. 
% The data and annotations can be found through the provided link.

Multimodal Large Language Models (MLLMs) have demonstrated proficiency in handling 
%multi-modal tasks, but falter in precise visual reasoning. Current benchmarks predominantly evaluate MLLMs on single image reasoning, focusing on static object knowledge. However, reasoning over image sequences, crucial for real-world applications, remains a significant challenge. 
a variety of visual-language tasks. However, current MLLM benchmarks are predominantly designed to evaluate reasoning based on static information about a single image, and the ability of modern MLLMs to extrapolate from image sequences, which is essential for understanding our ever-changing world, has been less investigated. To address this challenge, this paper introduces \ourset, a new benchmark designed to assess MLLMs' sequential image reasoning abilities. \ourset features 4,761 diverse image sequences with varying lengths. We also employ a GPT-4 assisted method to evaluate MLLM reasoning performance. 
% Evaluation of 9 recent MLLMs on \ourset, including GPT-4V and Gemini, reveals their limited reasoning capabilities with sequential images. Absent external information, these models struggle to describe image sequences accurately, often leading to object and behavioral hallucinations/misrepresentations. 
Through a careful evaluation of nine recent MLLMs on \ourset, including GPT-4V and Gemini, we find that they struggle to accurately describe dynamic information about given image sequences, often leading to hallucinations/misrepresentations of objects and their corresponding behaviors.
% \mbc{previous sentence a bit vague so let's mention some concrete new kinds of errors/hallucinations/misrepresentations we found for image sequences} 
Our quantitative analysis and case studies identify three key factors impacting MLLMs' sequential image reasoning: the correlation between object and behavioral hallucinations, the influence of co-occurring behaviors, and the compounding impact of behavioral hallucinations.
Our dataset is available at \href{https://github.com/umd-huang-lab/Mementos}{https://github.com/umd-huang-lab/Mementos}.
% \mbc{make the last data release part a separate sentence and add github link in footnote}
% link
\end{abstract}

\section{Introduction}

The recent emergence of Multimodal Large Language Models (MLLMs) such as GPT-4V~\cite{openai2023gpt4v} and Gemini~\cite{geminiteam2023gemini} has shown strong visual-language understanding and generation capabilities in many areas, like image captioning and visual question answering.
Despite the notable performance of existing MLLMs, they often suffer from hallucination (a phenomenon where MLLMs produce inaccurate descriptions of the given images) due to insufficient reasoning capabilities, generating inaccurate responses in visual inference~\cite{liu2023hallusionbench, yue2023mmmu}. 
Thus, monitoring the reasoning capability is of great importance in understanding the ability and the limitations of MLLMs and applying MLLMs in the real world.
Previous benchmarks, such as those in \citet{liu2023hallusionbench} and \citet{yue2023mmmu}, have primarily addressed evaluating reasoning in each individual image, relying on static and object-centric knowledge.
However, they are insufficient to comprehensively assess the reasoning capabilities of MLLMs due to a lack of time-varying object behaviors or events.

\begin{figure*}[!h]
\centering
\includegraphics[width=\textwidth]{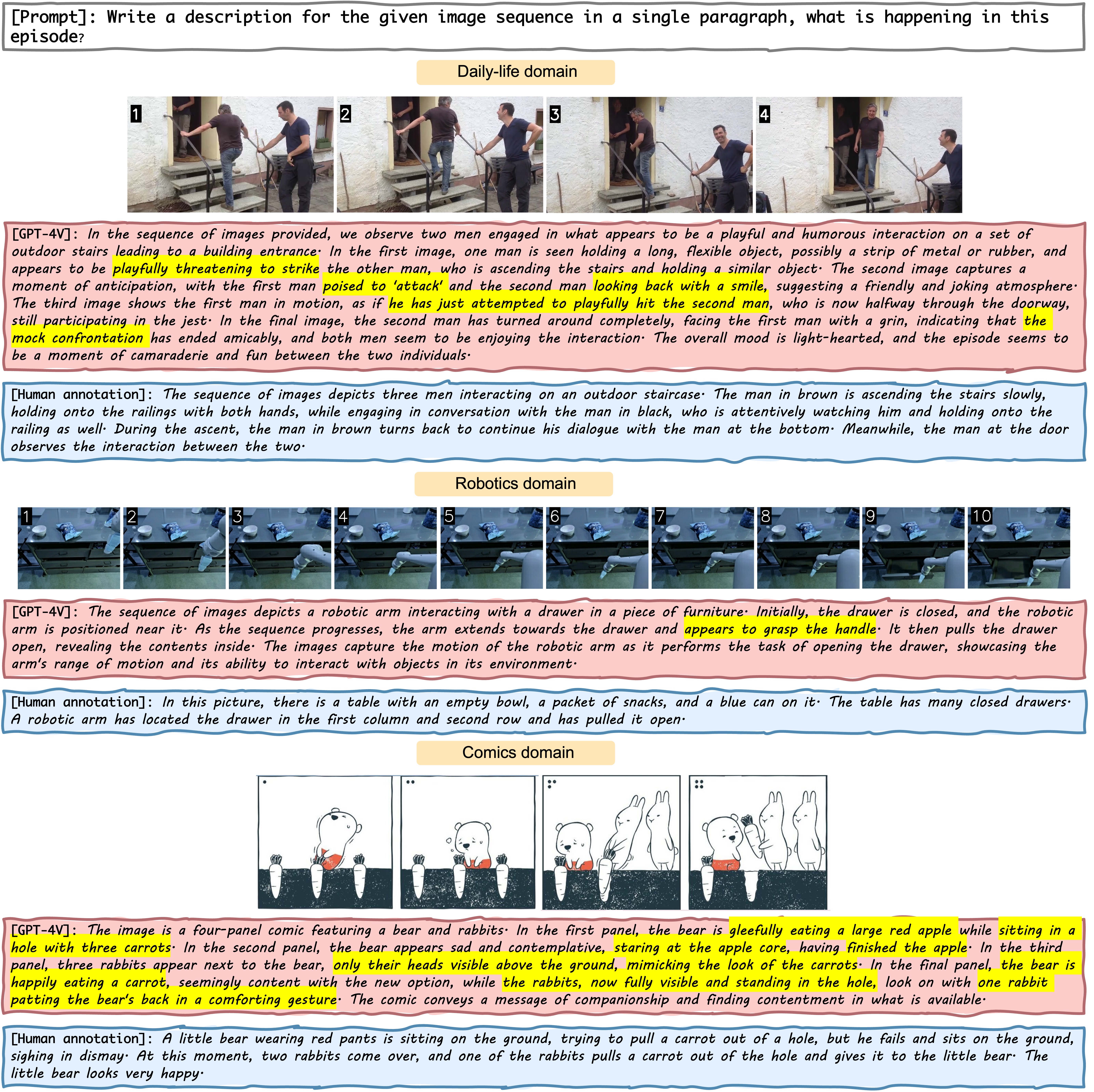}
\vspace{-18pt}
\caption{Examples of hallucinations by GPT-4V in three domains on \ourset. The red box shows the description generated by GPT-4V based on the given prompt, and the human-annotated descriptions are in the blue box. Texts highlighted in yellow are hallucination parts generated by GPT-4V. This illustrates that even GPT-4V experiences severe hallucinations when reasoning from image sequences.}
\vspace{-18pt}
\label{fig:Demo}
\end{figure*}

To investigate the capabilities of Multi-Modal Language Models (MLLMs) in dynamic reasoning across image sequences, we present a new benchmark, \textit{\ourset}. 
This benchmark focuses on the complex task of monitoring and deciphering the \textit{positional changes} of objects within an image sequence, followed by the inference of behavioral patterns and logical connections among them. 
Such an endeavor requires the interpretation of the overarching context based on time-variant visual elements, posing a greater challenge than the analysis of static scenes. Concretely, \ourset consists of 4,761 image sequences with varying episode lengths\footnote{Here, an episode refers to a specific event or series of events depicted in the image sequence.}, encompassing diverse scenarios from everyday life, robotics tasks, and comic-style storyboards.  Each sequence is paired with a human-annotated description of the primary objects and their behaviors within the sequence.

To assess the reasoning capability of MLLMs on \ourset, we employ a \textit{GPT-4}-assisted evaluation procedure:
after an MLLM produces a description for an image sequence, 
we extract behavior and object keywords from both AI-generated and human-annotated descriptions using GPT-4. We then use keyword matching to assess the degree of behavioral and object hallucinations.
To refine the correctness of this evaluation, we have developed behavior and object synonym graphs for each domain. These graphs facilitate more precise keyword matching, ensuring a thorough and nuanced analysis of the MLLMs' reasoning abilities.
Besides, we also provide the comparison with human evaluation to demonstrates that the GPT-4-assisted evaluation procedure is very reliable.

We evaluated the reasoning proficiency of \textit{nine leading-edge MLLMs} on \ourset, encompassing both black-box and open-source models. 
Our findings indicate that \ourset poses a considerable challenge to these current MLLMs. 
For instance, as depicted in Figure~\ref{fig:Demo}, GPT-4V exhibits notable behavioral and object hallucinations in various domains during image sequence reasoning. 
Behavioral hallucinations are defined as the MLLMs' erroneous interpretations or predictions of entity actions, while object hallucinations pertain to the inaccurate identification or creation of objects within the image sequences. 
Notably, behavioral hallucinations were more frequent than object hallucinations, highlighting a significant deficiency in MLLMs' capability to deduce events from image sequences.

Furthermore, our research pinpoints three principal factors that lead to the reasoning failures of MLLMs: (1) the interconnectedness of object and behavioral hallucinations, (2) the impact of co-occurring behaviors, and (3) the cumulative effect of behavioral hallucinations. The objective of our proposed benchmark and analyses is to shed light on innovative approaches to augment the reasoning abilities of MLLMs and to reduce hallucinations in their subsequent advancements.

\section{\ourset}
In this section, we introduce \ourset, a novel and challenging benchmark designed to test the reasoning capability of Multimodal Large Language Model (MLLM) under sequential image input. 
Initially, we detail the data gathering and annotation methodology for \ourset, alongside an overview of its data distribution. 
Subsequently, we outline the evaluation protocol and the metric employed to assess the reasoning prowess of MLLMs in \ourset.

\subsection{\ourset Benchmark}
\subsubsection{Dataset Composition}
\ourset comprises 4,761 image sequences of varying lengths, predominantly sourced from Daily-life, Robotics, and Comics domains. Detailed statistics are provided in Table~\ref{tab: data_stat}. This diverse collection is instrumental in evaluating the comprehensive time-varying reasoning abilities of MLLMs. Specifically, the robotics data, closely associated with embodied AI or real-world contexts, and the comic-style storyboard data, rich in stylistic and episodic diversity in image sequences, significantly enhance the benchmark's relevance and robustness.

% \vspace{-5pt}
\begin{table}[!htb]
\small
\setlength{\abovecaptionskip}{0.2cm}
\centering
\caption{The number of image sequences in different categories within \ourset. 
}
\begin{tabular}{l|c|c|c}
\toprule 
 & Total & Train Set & Val set \\
\midrule
Daily-life & 3505 & 3055 & 450   \\
Robotics & 1101 & 902 & 199 \\
Comics & 155 & 105 & 50   \\
\bottomrule
\end{tabular}
\label{tab: data_stat}
\vspace{-15pt}
\end{table}
% \vspace{-15pt}

\paragraph{Daily-life} 
The Daily-life image sequences in \ourset are derived from video clips in the Next-QA dataset, as cited in \citet{Xiao_2021_CVPR}. These sequences represent a range of everyday life scenarios. We have selectively extracted videos from the Next-QA Training set, specifically those with frame counts ranging from 400 to 2,500.
To balance the challenge of testing MLLMs' reasoning capabilities against the risk of losing critical information, our methodology involves retaining the first frame of each video. Subsequently, we sample one image every 100 frames. The collected images from this sampling process then form an image sequence that corresponds to the original video. This approach ensures a rigorous yet feasible evaluation of MLLMs' reasoning abilities in dynamically evolving everyday scenarios.

\paragraph{Robotics} 
For the Robotics data, we utilized videos from various sub-datasets within Open X-Embodiment~\cite{open_x_embodiment_rt_x_2023}. 
Open X-Embodiment aggregates video datasets from multiple university laboratories, showcasing a variety of tasks performed by different robotic systems.
We meticulously selected sub-datasets from Open X-Embodiment that offer video resolutions exceeding 128x128 and exhibit a high degree of task diversity. From these chosen sub-datasets, a total of 1,101 videos were sampled. The precise number of videos sourced from each sub-dataset is detailed in Appendix~\ref{app_rtx}.
For video sampling, our approach varied based on the length of the videos. Videos exceeding 100 frames were processed by sampling one image every $n/20$ frames, where $n$ represents the total frame count of the video. Conversely, for videos with frame counts ranging from 20 to 100, we sampled one image every 5 frames. This ensures the formation of comprehensive and representative image sequences for each video, catering to the evaluation of MLLMs in diverse and complex robotic contexts.

\paragraph{Comics} 
The Comics data is composed of wordless multi-panel comics of diverse styles, curated from online sources. 
Unlike Daily-life and Robotics sections, where image sequences are uniformly extracted from video frames, the comics represent intentionally selected key moments within a narrative, manually illustrated by artists.
This distinction sets our dataset apart from conventional video datasets. 
In addition to traditional comics, this category also incorporates 20 storyboards from movies reimagined in comic style.
We have further deconstructed these comics into individual image sequences by taking screenshots.
This approach enables a unique exploration of sequential visual reasoning, enhancing the diversity and complexity of the dataset for evaluating MLLMs.

\subsubsection{Dataset Annotation}

For each image sequence in \ourset, we have meticulously annotated a ground truth description that captures the unfolding events. 
These descriptions focus on the primary objects and their respective behaviors, where \textit{behavior} refers to a verb or verb phrase associated with the object in question.

For the Daily-life data, we initially employed GPT-4V(ision)~\cite{openai2023gpt4}, to amalgamate and reformulate the questions and answers from the Next-QA videos into single paragraph descriptions. 
This method significantly expedited the manual annotation process. Following this, we conducted a thorough manual review of these automated descriptions, making necessary adjustments. 
This included rectifying inaccuracies, removing non-existent episodes, and adding missing details to ensure alignment with the actual image sequences. 
To ensure reliability, we implemented a cross-validation step, where a separate set of annotators performed a secondary review.
For the Robotics and Comics categories, the annotation process was entirely manual, conducted by human annotators. These annotations were then subjected to a verification process by the authors which ensures the accuracy and consistency of the descriptions across all categories.
% providing a robust foundation for evaluating the MLLMs' reasoning capabilities.

\subsubsection{Dataset Statistics}

In showcasing the extensive diversity of \ourset, we present a detailed overview of the data distribution within the \ourset validation set. 
Our analysis focuses on two key dimensions: the length of the image sequence and the length of the episode. 
The length of an image sequence is defined by the number of frames it contains, while the episode length is determined by the total number of events depicted in the sequence.
A longer image sequence necessitates the MLLM to process a larger number of images, thereby challenging the model's capacity to manage sequences spanning broader timeframes. 
A greater episode length signifies that the image sequence encompasses more intricate scenarios. 
% This complexity demands that the MLLM demonstrate enhanced reasoning abilities to effectively interpret and understand these detailed and elaborate sequences. 
% This dual-perspective analysis underscores the dataset's capability to rigorously test and refine the reasoning prowess of MLLMs in diverse and dynamic visual contexts.

\begin{figure}[!htbp]
\centering
% \vspace{-15pt}
\includegraphics[width=0.45\textwidth]{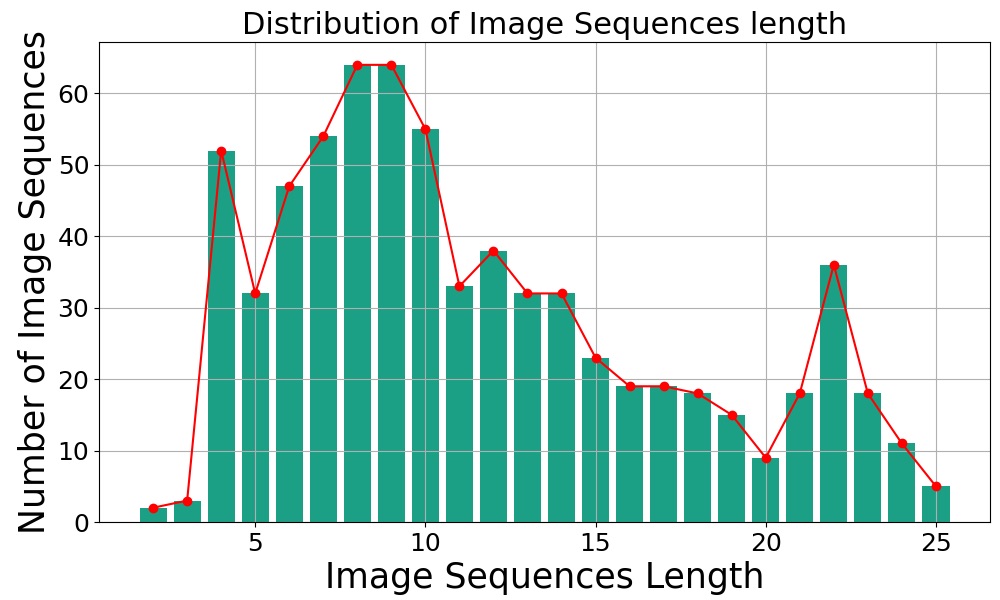}
% \vspace{-10pt}
\caption{Distribution of image sequence length in \ourset Val set.}
\label{fig:stat_is_len}
\end{figure}

\begin{figure}[!htbp]
\centering
% \vspace{-20pt}
\includegraphics[width=0.45\textwidth]{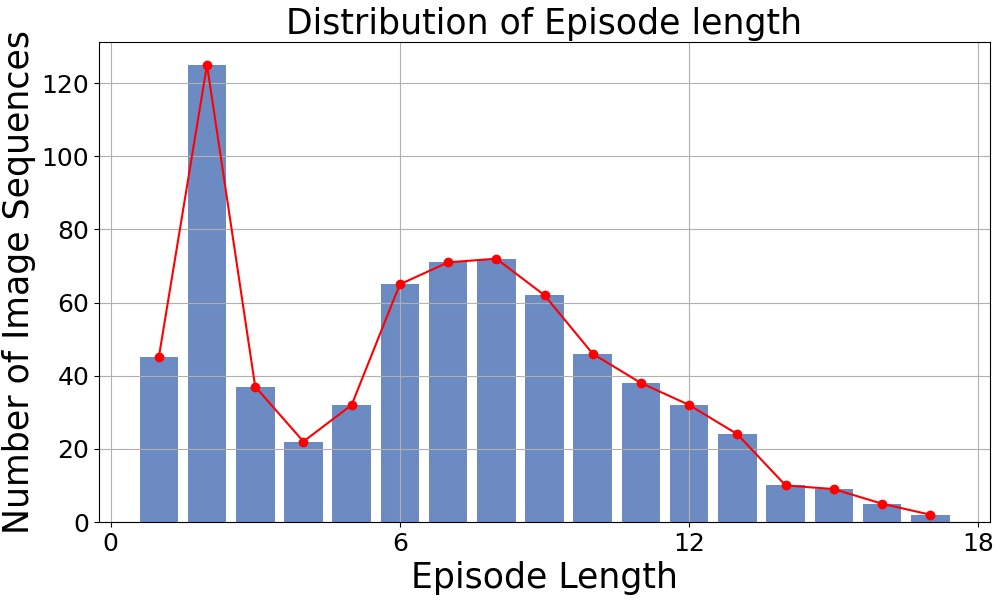}
% \vspace{-10pt}
\caption{Distribution of episode length in \ourset Val set.}
\label{fig:stat_ep_len}
\vspace{-15pt}
\end{figure}
% \vspace{-20pt}

\paragraph{Image sequence length}
For the image sequence length, we count the number of frames in each image sequence. As shown in Figure~\ref{fig:stat_is_len}, 
the majority of image sequences are between 4 and 14 frames in length. 
67.38\% of image sequences contain 4 to 14 frames, yet 31.90\% of sequences are composed of longer frames - more than 15 frames.

\paragraph{Episode length}
To quantify the episode length within each image sequence of \ourset, we employed GPT-4 for extracting behavior keywords, specifically verbs associated with objects, from the human-annotated descriptions. This extraction was facilitated using a pre-defined manual prompt, details of which can be found in Appendix~\ref{app_prompt}. Following the extraction, we calculated the length of the behavior list for each image sequence.
A lengthier behavior list signifies a more extended episode within the image sequence, which inherently poses a greater challenge for the MLLM in comprehending the entire image sequence. As illustrated in Figure~\ref{fig:stat_ep_len}, a significant portion of the image sequences, particularly those from the robotics data, feature episode lengths ranging between 1 and 3. This is mainly attributed to the dominance of two-action episodes like `pick up and place', `move and pull open', `locate and push'. Meanwhile, the remaining data exhibits a normal distribution for episode lengths spanning 4 to 17.

\begin{figure}[!htbp]
\centering
% \vspace{-15pt}
\includegraphics[width=0.5\textwidth]{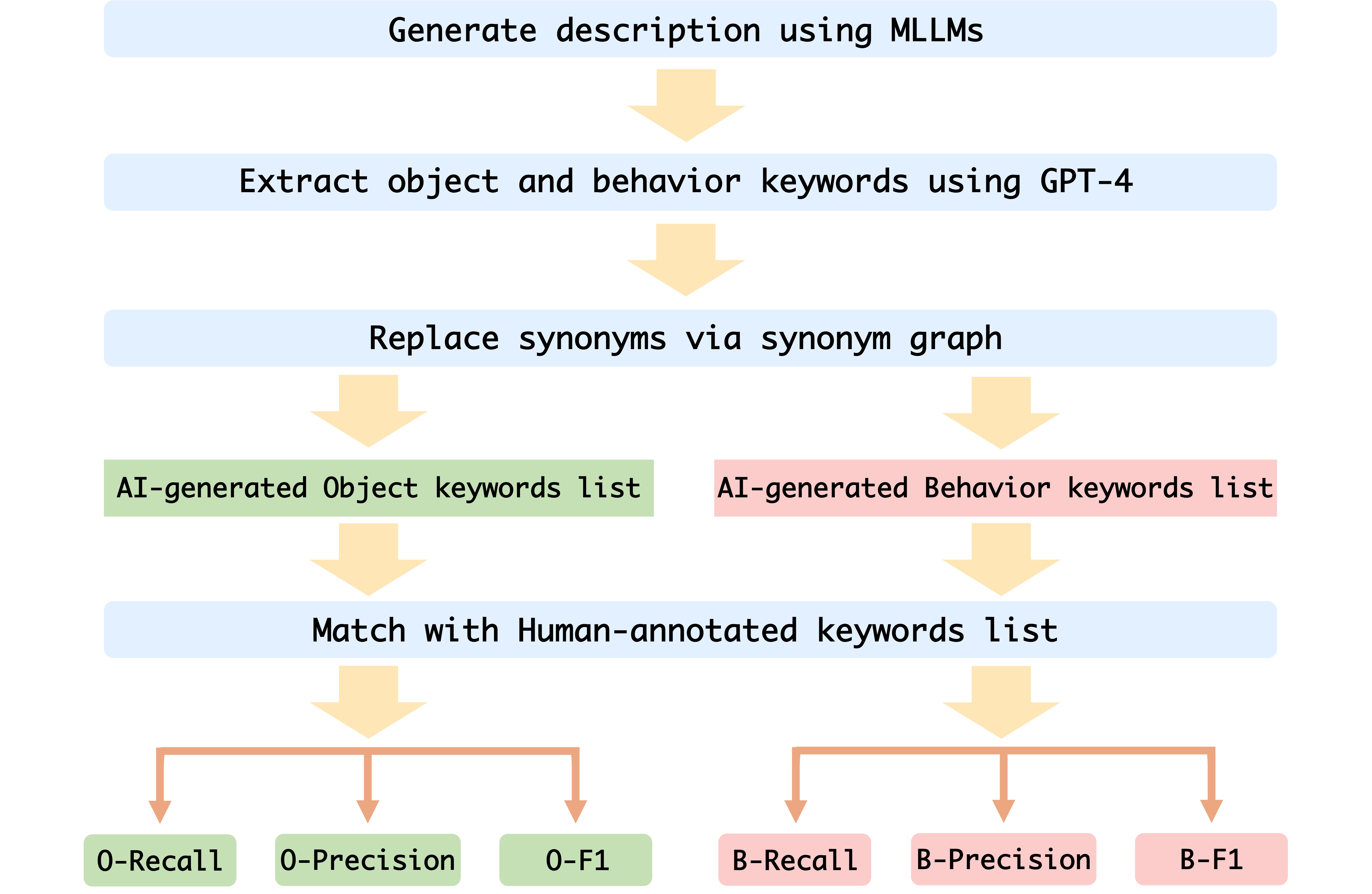}
% \vspace{-20pt}
\caption{GPT-4-assisted evaluation procedure. We employ "O-" and "B-" to indicate objects and behaviors, respectively.}
\vspace{-15pt}
\label{fig:proced}
\end{figure}

\subsection{Evaluation Procedure and Metrics}

In this section, we illustrate how to evaluate the descriptions generated by MLLMs, including the evaluation procedure and metrics. 
\paragraph{Procedure} As shown in Figure~\ref{fig:proced}, we use an image sequence and a pre-designed prompt together as the input for MLLMs, and generate the description aligned with the corresponding image sequence. 
Next,
we ask GPT-4 to extract object and behavior keywords using the AI-generated description. 
We then match the obtained keywords with the \textit{synonym graph} we built, replacing the matched keywords with the root word from the synonym graph. 
Finally, we obtain two lists of keywords: AI-generated object list and AI-generated behavior list.
We note that the proposed keyword extraction leveraging GPT-4 is surprisingly reliable and accurate, which is competitive with human extraction.
Please refer to Appendix~\ref{app: huamneval} for a detailed comparison.

\paragraph{Synonym graph} The synonym graph is an unilateral digraph where each edge connects two nodes representing words or phrases. For instance, given a synonym pair (pick up, lift up), an edge would be directed from `lift up' to `pick up'.
In each synonym pair, the first word, originating from the human-annotated keyword list, is referred to as the root word, while the second word comes from the AI-generated keyword list.
To construct this synonym graph, we first use GPT-4 to extract object and behavior keywords from all human-annotated descriptions in the Val set, forming a human-annotated keyword list. 
Then, we generate descriptions on the Val set using GPT-4V, LLAVA, and Gemini and use GPT-4 to extract object and behavior keywords. 
After that, 
we manually match these words with the human-annotated keyword list to identify all synonym pairs and add them as edges to the synonym graph.
Given a word or phrase, this synonym graph can quickly match the corresponding root word if a synonym exists in the human-annotated keyword list, completing the keyword replacement.
For convenience in evaluation, we maintain separate synonym graphs for objects and behaviors of different categories.
We make all constructed synonym graphs publicly available as open-source resources.

\paragraph{Metrics} After obtaining the AI-generated object list and AI-generated behavior list for each image sequence, we utilize the corresponding human-annotated object list and human-annotated behavior list as the ground truth to calculate `Recall,' `Precision,' and `F1 metrics' at both the object and behavior levels. 
These metrics are used to measure the understanding capabilities of different MLLMs regarding the image sequence episode.
`Recall' reflects the accuracy of an MLLM's reasoning about episodes in an image sequence, while `precision' focuses on assessing the severity of hallucinations that occur when understanding the image sequence.
\section{Experiments}

In our experimental section, we delve into two key questions:
(a) We examine the reasoning \textbf{capabilities} of current black-box and open-source MLLMs on \ourset. Specifically, we assess the \textbf{severity} of object and behavioral hallucinations in these models.
(b) We investigate the underlying \textbf{causes} of reasoning failures in MLLMs when interpreting image sequences.

\subsection{Baseline evaluation}
\subsubsection{Models} 
We establish our baseline using 9 popular black-box and open-source MLLMs.
The black-box MLLMs include GPT-4V~\cite{openai2023gpt4} and Gemini~\cite{geminiteam2023gemini}, and the open-source MLLMs are Video-LLaMA-2~\cite{damonlpsg2023videollama}, Chat-UniVi~\cite{jin2023chatunivi}, LLaVA-1.5~\cite{liu2023improved}, MiniGPT4~\cite{zhu2023minigpt}, MiniGPT5~\cite{zheng2023minigpt5}, mPLUG\_Owl-v2~\cite{ye2023mplug}, and InstructBLIP~\cite{dai2023instructblip}.
Additionally, considering that only a few open-source MLLMs are designed to process sequential images or videos (Video-LLaMA-2 and Chat-UniVi), we adapt input for other open-source MLLMs by combining all frames from an image sequence into one composite image. This input is referred to as the combined-input (c-input) setting. For black-box MLLMs and Chat-UniVi, we conduct evaluations using both the c-input setting and an alternative approach where frames from the image sequence are input sequentially, termed the sequential-input (s-input) setting. For Video-LLaMA-2, we only test the performance on s-input setting.

\begin{figure}[!htbp]
\centering
\vspace{-5pt}
\includegraphics[width=0.45\textwidth]{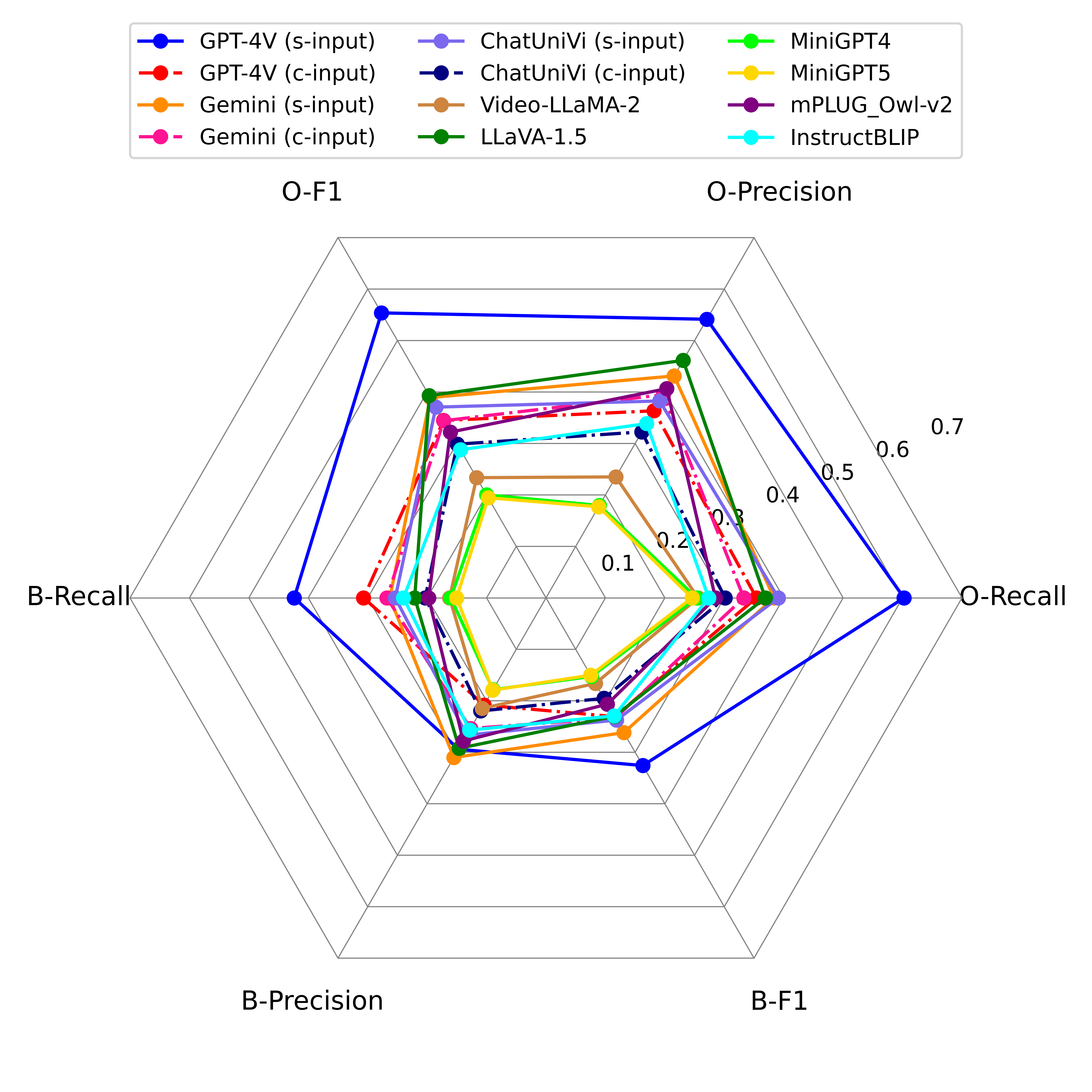}
% \vspace{-10pt}
\caption{Comparison of the six metrics for different MLLMs on \ourset.}
\label{fig:total_exp}
\vspace{-15pt}
\end{figure}
% \vspace{-15pt}

\subsubsection{Evaluation results}
We evaluate all baseline MLLMs on \ourset and report the results in Figure~\ref{fig:total_exp}.
Besides, we provide the performance comparison of each baseline method across the three different domains (Daily-life, Robotics, and Comics) in Table~\ref{tab: eval_results}. We summarize our findings as follows:

\begin{table*}[!htb]
\small
\setlength{\abovecaptionskip}{0.2cm}
\centering
\caption{Evaluation of different MLLMs on \ourset.}
\vspace{-5pt}
\begin{tabular}{c|c|c|ccc|ccc}
\toprule 
\multirow{2}{*}{\textbf{Domain}} & \multirow{2}{*}{\textbf{Input type}} & \multirow{2}{*}{\textbf{Model}}&       & \textbf{Object}     &          &       & \textbf{Behavior} &  \\  
            &       &          & Recall & Precision & F1 & Recall & Precision & F1\\
\midrule
            & \multirow{4}{*}{Sequential}   &  GPT-4V  & \textbf{59.80\%} & \textbf{50.96\%} & \textbf{53.51\%} & \textbf{36.71\%} & 32.97\% & \textbf{33.59\%}  \\
            &              &  Gemini  & 35.92\% & 42.06\% & 37.10\% & 18.80\% & 29.42\% & 21.64\%  \\
            &              &  Video-LLaMA-2  & 31.59\% & 30.01\% & 29.37\% & 17.05\% & 28.19\% & 20.12\%  \\
            &              &  Chat-UniVi  & 40.74\% & 40.78\% & 39.13\% & 22.30\% & 31.10\% & 24.90\%  \\
\cmidrule{2-9}
            &       \multirow{8}{*}{Combined}        &  GPT-4V  & 39.45\% & 39.64\% & 38.04\% & 26.43\% & 23.59\% & 23.98\% \\
Daily-life  &              &  Gemini  & 31.17\% & 37.39\% & 32.38\% & 17.71\% & 25.65\% & 19.74\% \\
            &              &  Chat-UniVi  & 36.19\% & 38.88\% & 36.02\% & 21.80\% & 28.52\% & 23.73\%  \\
            &              &  LLaVa-1.5 & 37.72\% & 47.01\% & 40.18\% & 22.17\% & \textbf{37.33\%} & 26.65\% \\
            &      &  MiniGPT4  & 32.25\% & 23.14\% & 25.75\% & 18.09\% & 24.16\% & 19.45\% \\
            &              &  MiniGPT5  & 31.39\% & 22.62\% & 24.91\% & 18.42\% & 24.56\% & 19.85\% \\
            &              &  mPLUG\_Owl-v2 & 32.59\% & 47.17\% & 37.04\% & 17.96\% & 33.57\% & 22.13\% \\
            &              &  InstructBLIP & 31.82\% & 41,14\% & 34.28\% & 22.40\% & 30.30\% & 24.55\% \\
\midrule
\midrule
            & \multirow{4}{*}{Sequential}   &  GPT-4V  & \textbf{63.94\%} & \textbf{65.42\%} & \textbf{62.99\%} & \textbf{60.72\%} & 24.43\% & 33.95\%  \\
            &              &  Gemini  & 43.80\% & 46.26\% & 43.15\% & 46.43\% & \textbf{38.13\%} & \textbf{39.38\%}  \\
            &              &  Video-LLaMA-2  & 13.41\% & 10.33\% & 11.15\% & 17.04\% & 8.96\% & 11.23\%  \\
            &              &  Chat-UniVi  & 35.40\% & 32.57\% & 32.39\% & 32.24\% & 16.69\% & 21.14\%  \\
\cmidrule{2-9}
            &      \multirow{8}{*}{Combined}        &  GPT-4V  & 27.87\% & 31.86\% & 28.58\% & 44.72\% & 16.54\% & 23.58\% \\
Robotics    &              &  Gemini  & 34.78\% & 41.66\% & 36.16\% & 47.29\% & 29.59\% & 34.17\% \\
            &              &  Chat-UniVi  & 17.74\% & 18.32\% & 17.07\% & 19.81\% & 10.01\% & 12.54\%  \\
            &              &  LLaVa-1.5 & 36.88\% & 46.62\% & 39.31\% & 25.27\% & 14.80\% & 17.95\% \\
            &      &  MiniGPT4  & 10.97\% & 7.28\% & 8.16\% & 13.40\% & 5.88\%  & 7.76\%  \\
            &              &  MiniGPT5  & 9.75\% & 6.52\% & 7.16\% & 8.96\% & 4.53\% & 5.43\% \\
            &              &  mPLUG\_Owl-v2 & 19.75\% & 26.70\% & 21.99\% & 26.46\% & 16.59\% & 19.51\% \\
            &              &  InstructBLIP & 17.96\% & 18.65\% & 17.29\% & 31.41\% & 19.08\% & 22.69\% \\
\midrule
\midrule
            & \multirow{4}{*}{Sequential}   &  GPT-4V  & \textbf{49.53\%} & 37.57\% & \textbf{41.71\%} & \textbf{19.97\%} & 17.29\% & \textbf{18.11\%}  \\
            &              &  Gemini  & 38.57\% & 40.64\% & 38.53\% & 15.23\% & 19.11\% & 16.30\%  \\
            &              &  Video-LLaMA-2  & 20.26\% & 17.59\% & 18.09\% & 5.45\% & 11.07\% & 6.81\%  \\
            &              &  Chat-UniVi  & 28.04\% & 31.61\% & 28.13\% & 10.42\% & 15.74\% & 11.97\%  \\
\cmidrule{2-9}
            &        \multirow{8}{*}{Combined}      &  GPT-4V  & 29.23\% & 24.64\% & 25.90\% & 13.19\% & 13.09\% & 12.90\% \\
Comics      &              &  Gemini  & 41.25\% & \textbf{45.07\%} & 41.18\% & 15.37\% & \textbf{20.55\%} & 16.42\% \\
            &              &  Chat-UniVi  & 25.12\% & 28.08\% & 25.51\% & 8.85\% & 10.67\% & 9.31\%  \\
            &              &  LLaVa-1.5 & 29.44\% & 35.61\% & 30.97\% & 8.63\% & 13.56\% & 10.27\% \\
            &              &  MiniGPT4  & 20.50\% & 13.94\% & 15.74\% & 7.95\% & 8.64\% & 7.98\% \\
            &              &  MiniGPT5  & 22.94\% & 18.11\% & 19.42\% & 8.88\% & 11.92\% & 9.94\% \\
            &              &  mPLUG\_Owl-v2 & 26.82\% & 37.74\% & 29.49\% & 8.70\% & 20.85\% & 11.74\% \\
            &              &  InstructBLIP & 25.02\% & 29.15\% & 25.10\% & 8.25\% & 10.48\% & 8.97\% \\
\bottomrule
\end{tabular}
\label{tab: eval_results}
\vspace{-20pt}
\end{table*}

\paragraph{GPT-4V (s-input) and LLaVA-1.5 are the best-performing models among black-box and open-source MLLMs, respectively.}
As shown in Figure~\ref{fig:total_exp}, except for being on par with Gemini (s-input) and LLaVA-1.5 in behavior precision, GPT-4V with s-input demonstrates the best reasoning capability compared with all other MLLMs in understanding image sequences. 
Among open-source models, LLaVA1.5 performs the best, nearly matching or even surpassing the black-box model Gemini in object comprehension, but its ability to infer behaviors from image sequences is weaker compared to Gemini and GPT-4V. 
Although Video-LLaMA-2 and Chat-UniVi are designed for video understanding, they do not show an advantage over LLaVA-1.5, especially Video-LLaMA-2, which performs notably worse compared to LLaVA-1.5.
The weakest models in understanding image sequences are MiniGPT4 and MiniGPT5, with a significant gap in every metric compared to the other baselines.
It's noteworthy that under c-input setting, the performance of black-box MLLMs does not significantly differ from that of open-source MLLMs. 
LLaVA-1.5 and mPLUG\_Owl-v2 meet or even exceed the black-box MLLMs on many metrics.

\paragraph{MLLMs possess a much stronger ability on reasoning objects in image sequences than they do on reasoning behaviors.}
We find that all MLLM methods perform significantly better on the three metrics for objects than those for behaviors. 
Taking the best-performing GPT-4V as an example, it achieves over 50\% on all three object metrics, with recall even reaching 60\%, indicating it can effectively recognize the main objects in an image sequence. 
However, for behaviors, GPT-4V scores only around 30\%, with the best recall metric barely exceeding 40\%. 
Despite this, GPT-4V is still the best-performing MLLM in reasoning behaviors. 
This suggests that current MLLMs do not possess strong abilities to autonomously infer the behaviors from given sequential images, indicating the importance of our benchmark in highlighting the limitations in the reasoning abilities of MLLMs.

\paragraph{Reasoning capability of MLLMs varies across different domains.}
From Table~\ref{tab: eval_results}, 
we find that black-box models perform best in the robotics domain across the three domains, while open-source models show relatively better performance in the daily-life domain.
% \gb{This is an interesting yet very strange/unexpected finding. Why would GPT4V perform better on robotics data than on daily-life data if it's trained on large-scale Internet data?} 
Analyzing each domain specifically, it is evident that in the daily-life domain, the performance of all methods, except for GPT-4V (s-input), does not vary significantly. 
The main reason for the performance gap between open-source MLLMs and black-box MLLMs is the noticeably lower metrics of open-source models compared to black-box models in the robotics and comics domains. 
The recall, precision, and F1 of both object and behavior for black-box MLLMs are almost more than double those of open-source models. 
We speculate that one reason for this phenomenon is the distribution shift between \ourset and the training data of open-source MLLMs. 
The limitations of the training data lead to weaker reasoning capability of open-source MLLMs. 

\subsection{Analysis of Failure Reasoning}
\label{sec: error_ana}
\begin{figure}[!htbp]
\centering
% \vspace{-15pt}
\includegraphics[width=0.45\textwidth]{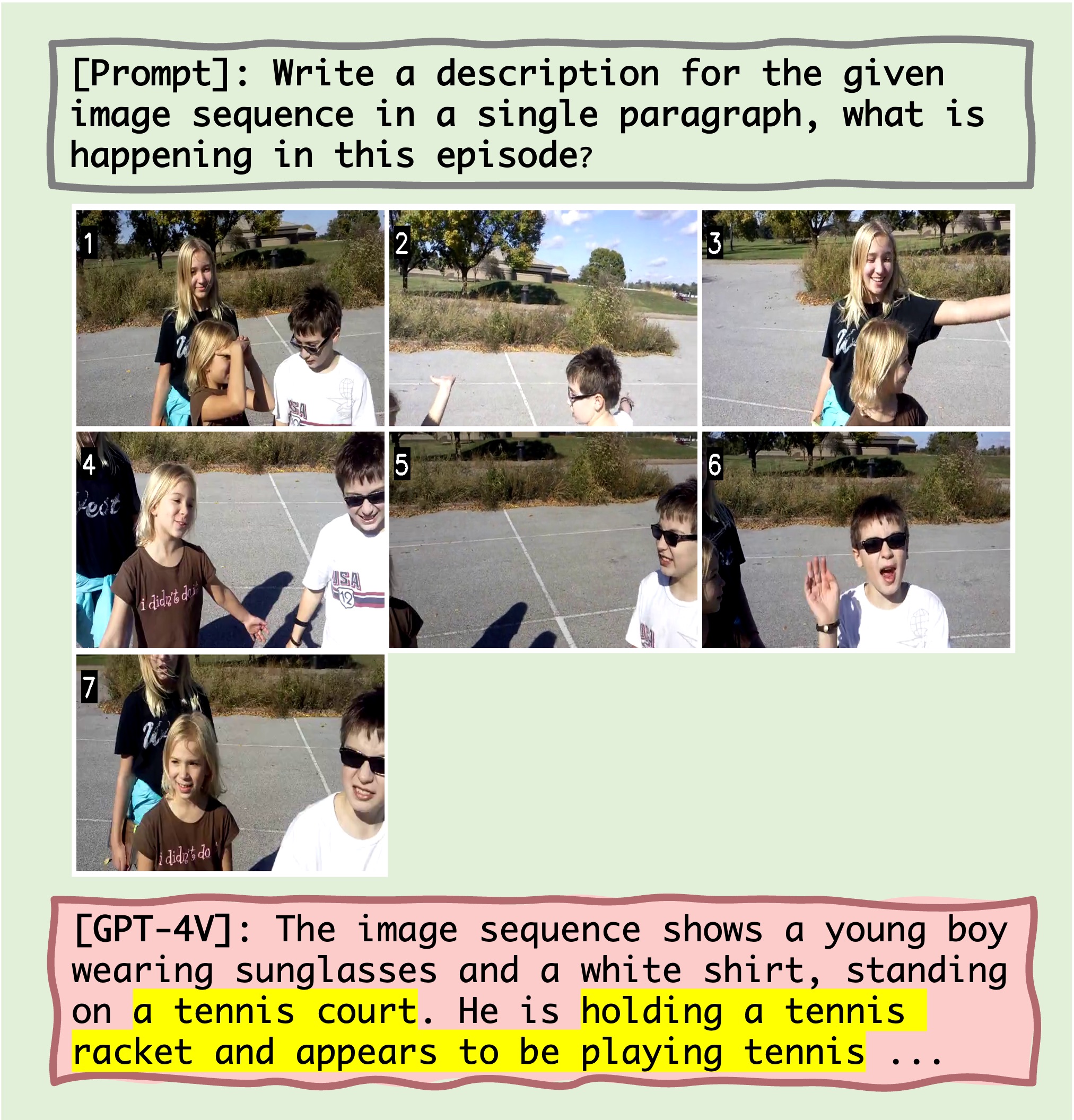}
% \vspace{-8pt}
\caption{A sample of failure reasoning case in Daily-life domain. The failure reason is object hallucination, correlation between object hallucination and behavioral hallucination, and co-occurrence behavior. Following the object hallucination of \textbf{\textit{tennis court}}, the LVLM subsequently exhibits behavioral hallucinations of \textbf{\textit{holding a tennis racket}} (correlation between object hallucination and behavioral hallucination) and \textbf{\textit{appears to be playing tennis}} (co-occurrence behavior).}
\vspace{-15pt}
\label{fig:case_mainpaper}
\end{figure}

In this section, we will provide reasons for failure reasoning results in current MLLMs, combining specific quantitative analyses and case studies.
Since behavioral hallucination is a unique phenomenon in image sequence reasoning, and the causes of object hallucination are not significantly different from those in single image reasoning, we only present the reasons leading to behavioral hallucination in this paper.
Due to space limitations, please refer to the Appendix~\ref{app_case} for specific case studies. The following are our main findings:

\paragraph{Interplay between object and behavioral hallucinations in MLLMs.}
A key hypothesis underpinning behavioral hallucination is that incorrect object identification leads to subsequent inaccuracies in behavior identification. To test this, we evaluated the correlation coefficients between object and behavioral hallucinations across different domains for various MLLMs, as detailed in Table~\ref{tab: cor}.
Our findings reveal that, for most MLLMs, the correlation coefficients in the three domains fluctuate between 0.1 and 0.4, suggesting a weak yet present correlation. This outcome supports the hypothesis that object hallucination contributes to behavioral hallucination to some extent.
Case studies further reveal that after an object hallucination occurs, MLLMs tend to describe behaviors related to the hallucinated object, even if these behaviors do not exist in the image sequence. 
As shown in Figure~\ref{fig:case_mainpaper}, after recognizing a scene as a tennis court, a MLLM might describe a person playing tennis. 
Interestingly, in the robotics domain, there is a negligible correlation between object and behavioral hallucinations in black-box MLLMs. This divergence is likely because behaviors in robotics are predominantly linked to robotic arms, which these MLLMs generally identify correctly. 

\paragraph{The impact of co-occurrence on behavioral hallucination.}
In line with object hallucination phenomena, as noted in \citet{li2023evaluating} and \citet{zhou2023analyzing}, MLLMs demonstrate a tendency to generate behaviors that are commonly paired together when reasoning through image sequences. This proclivity exacerbates the problem of behavioral hallucination, especially in the field of robotics.
Consider the case in Figure~\ref{fig:Demo} where a robotic arm is tasked with opening a drawer by \textbf{\textit{grabbing its side}}. 
MLLMs might erroneously depict the sequence as the arm \textbf{\textit{grabbing the handle}} first, followed by pulling the drawer open, since \textbf{\textit{grabbing the handle}} is a more co-occurring behavior with `pull open'. 
Despite the final outcome being accurately described, such errors in key details are unacceptable in robotics.
This issue is of particular concern given the growing inclination to utilize MLLMs as reward functions in robotic training, as discussed in recent studies~\cite{{ma2023liv}, {sontakke2023roboclip}, {rocamonde2023vision}, {baumli2023vision}}.
Such subtle yet significant behavioral hallucinations can critically affect the quality of the reward function, leading to potential mislearning of behaviors in robotic systems.
For more detailed case studies, please refer to the Appendix~\ref{app_case}.

\begin{table}[!htb]
\setlength{\abovecaptionskip}{0.2cm}
\centering
\caption{Correlation coefficient between behavioral hallucination and object hallucination of different MLLMs on \ourset.} 
\resizebox{0.45\textwidth}{!}{ 
\begin{tabular}{c|c|c|ccc}
\toprule 
\textbf{Domain} & \textbf{Input type} & \textbf{Model} & Recall & Precision & F1 \\
\midrule
            &\multirow{4}{*}{Sequential}                & GPT-4V & 0.120 & 0.188 & 0.132   \\
            &                & Gemini & 0.165 & 0.179 &  0.146   \\
            &                & Video-LLaMA-2 & 0.197 & 0.067 &  0.125   \\
            &                & Chat-UniVi & 0.138 & 0.178 &  0.137   \\
\cmidrule{2-6}
            &\multirow{8}{*}{Combined}                 & GPT-4V& 0.242 & 0.182 &  0.199 \\
Daily-life  &                & Gemini  & 0.158 & 0.179 & 0.152   \\
            &                & Chat-UniVi & 0.127 & 0.184 &  0.172   \\
            &                & LLaVa-1.5 & 0.112 & 0.134 & 0.106  \\
            &                & MiniGPT4 & 0.135 & 0.145 &  0.115  \\
            &                & MiniGPT5 & 0.126 & 0.188 &  0.146  \\
            &                & mPLUG\_Owl-v2 & 0.106 & 0.113 &  0.069  \\
            &                & InstructBLIP & 0.133 & 0.125 &  0.127 \\
\midrule
\midrule
            &\multirow{4}{*}{Sequential}                & GPT-4V & -0.012  & 0.022  & 0.011   \\
            &                & Gemini & 0.027  & 0.144 &  0.101   \\
            &                & Video-LLaMA-2 & 0.107 & 0.107 &  0.109   \\
            &                & Chat-UniVi & 0.038 & 0.121 &  0.089   \\
\cmidrule{2-6}
            &\multirow{8}{*}{Combined}                 & GPT-4V& 0.041  & -0.022  & 0.008 \\
Robotics    &                & Gemini  & -0.049 & -0.086 & -0.106   \\
            &                & Chat-UniVi & 0.189 & 0.242 &  0.207   \\
            &                & LLaVa-1.5 & 0.135  & 0.123  & 0.157  \\
            &                & MiniGPT4 & 0.186 & 0.316 &  0.233  \\
            &                & MiniGPT5 & 0.056  & 0.027  &  0.045  \\
            &                & mPLUG\_Owl-v2 & 0.244 & 0.163 &  0.231  \\
            &                & InstructBLIP & 0.227 & 0.235 &  0.253 \\
\midrule
\midrule
            &\multirow{4}{*}{Sequential}                & GPT-4V & 0.045  & 0.225  & 0.158    \\
            &                & Gemini & 0.176  & 0.081 &  0.144    \\
            &                & Video-LLaMA-2 & 0.261 & 0.280 &  0.299   \\
            &                & Chat-UniVi & 0.239 & 0.331 &  0.221   \\
\cmidrule{2-6}
            &\multirow{8}{*}{Combined}                 & GPT-4V& 0.343 & 0.539  & 0.471 \\
Comics      &                & Gemini  & 0.187  & 0.121 &  0.167    \\
            &                & Chat-UniVi & 0.293 & 0.113 &  0.279   \\
            &                & LLaVa-1.5 & 0.062  & 0.101  & 0.088  \\
            &                & MiniGPT4 & 0.199  & 0.134  &  0.213  \\
            &                & MiniGPT5 & 0.324  & 0.366  &  0.339  \\
            &                & mPLUG\_Owl-v2 & 0.231 & -0.043 &  0.157  \\
            &                & InstructBLIP & 0.288 & 0.005 &  0.262 \\
\bottomrule
\end{tabular}
\label{tab: cor}
}
\vspace{-15pt}
\end{table}

\paragraph{The Snowball effect in behavioral hallucinations.}

\begin{figure}[!htbp]
\centering
\vspace{-15pt}
\subfigure[{Object}]{
\includegraphics[width=0.22\textwidth]{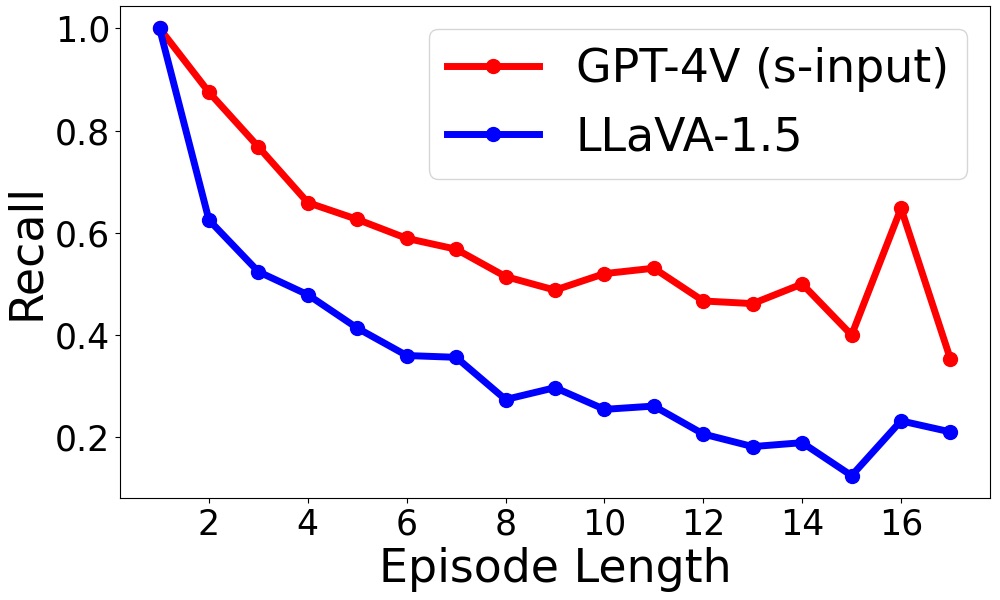}
\label{fig: o_re}
}
\hfil
\subfigure[{Behavior}]{
\includegraphics[width=0.22\textwidth]{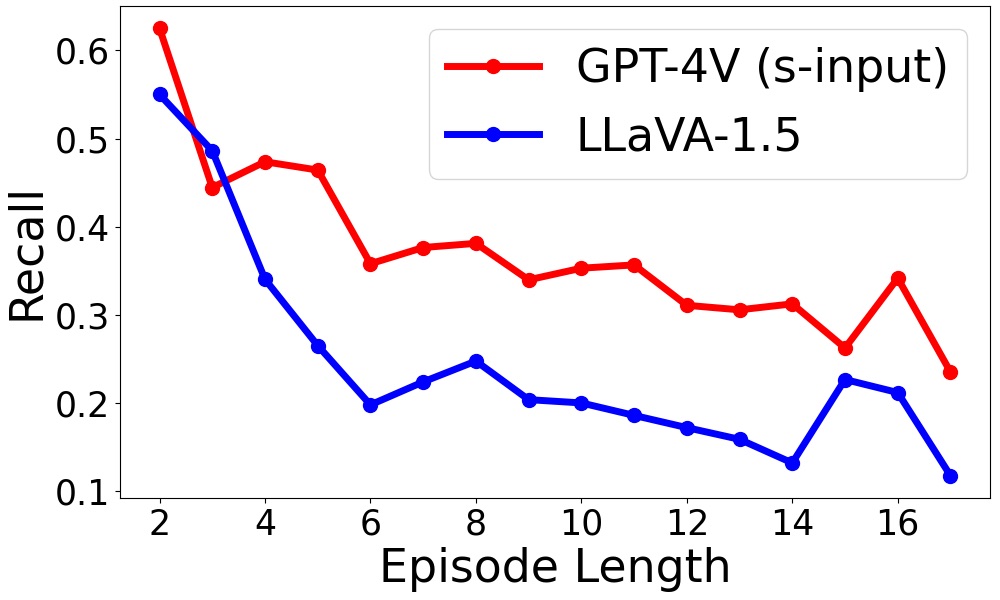}
\label{fig: a_re}
}
\vspace{-13pt}
\caption{The trend of changes in object and behavior recall for GPT-4V and LLaVA-1.5 in the Daily-life domain as the episode length of the image sequence increases.}
\label{fig: snowball}
\vspace{-15pt}
\end{figure}

In machine learning, the Snowball effect is a well-documented phenomenon, referring to the progressive accumulation or intensification of errors within a system, as discussed in \citet{{asadi2019combating}, {zhang2023language}, {wang2023coplanner}, {liu2023c}}. 
\citet{zhang2023language} notably highlight this phenomenon in Large Language Models.
Experiments on \ourset reveal that the snowball effect in both behavioral and object hallucinations becomes markedly pronounced when reasoning through image sequences. 
The temporal nature of image sequences, consisting of a series of frames rather than a solitary image, demands that MLLMs sequentially infer the narrative, frame by frame. This process makes the models susceptible to accumulating and exacerbating hallucinations if errors occur early in the sequence.
In our analysis, we specifically examined the trend of object and behavioral hallucination in GPT-4V and LLaVA-1.5 within the daily-life domain, correlating it with the increasing episode length of image sequences. As depicted in Figure~\ref{fig: snowball}, there is a noticeable decrease in object and behavior recall for both MLLMs as the episode length extends. This trend suggests a heightened susceptibility to hallucinations and a pronounced snowball effect in MLLMs when processing image sequences with a greater array of objects and behaviors.
Detailed case studies can be found in Appendix~\ref{app_case}.

\section{Related work}

\subsection{Benchmarking in MLLMs}
The advent of Multimodal Large Language Models (MLLMs) has prompted a reassessment of traditional benchmarks~\cite{lin2014microsoft, marino2019ok, hudson2019gqa}, originally conceived for Vision Language Models (VLMs). These existing benchmarks fail to sufficiently expose the robustness and hallucination issues inherent in MLLMs. Consequently, there is a growing impetus within the research community to devise more challenging benchmarks. This trend spans various domains, from question and answering (QA) reasoning~\cite{liu2023hallusionbench, yue2023mmmu}, to optical character recognition (OCR)~\cite{liu2023hidden}, and extends to the study of hallucinations~\cite{wang2023evaluation}, with benchmarks like POPE \cite{li2023evaluating} and Bingo \cite{cui2023holistic}. Additionally, comprehensive analyses of MLLMs, such as Mmbench \cite{liu2023mmbench}, Mm-vet\cite{yu2023mm}, LVLM-eHub\cite{xu2023lvlm}, SEED\cite{li2023seed}, GAVIE\cite{liu2023aligning}, and LAMM \cite{yin2023lamm}, are emerging.

Diverging from the focus on single images in prior studies, our paper introduces a novel benchmark that employs image sequences derived from video frames or comics, specifically examining the behavioral hallucinations in MLLMs. While \citet{chen2023autoeval} have employed a method of uniformly sampling frames from videos to create vision QA tasks for MLLMs, our benchmark is distinct in its challenge, as it prompts MLLMs to formulate descriptions of image sequences without the guidance of questions. This approach allows for a more nuanced evaluation of behavioral hallucinations and, by extension, a more precise assessment of MLLMs' reasoning capabilities.

\subsection{Hallucination in MLLMs}
Hallucinations in MLLMs, akin to those in Large Language Models (LLMs)~\cite{zhang2023siren, li2023halueval, zhou2023explore, chen2023hallucination}, represent a significant challenge. In MLLMs, hallucinations are characterized by inconsistencies between the model's output and the visual content~\cite{rohrbach2018object, wang2023evaluation}. Recent studies have explored various aspects of hallucination in MLLMs, covering topics such as object hallucination~\cite{li2023evaluating}, hallucination assessment in GPT-4V~\cite{cui2023holistic}, and knowledge hallucination~\cite{liu2023hallusionbench}.

While there are frameworks proposed for mitigating hallucinations~\cite{zhou2023analyzing, wang2023mitigating, leng2023mitigating, zhou2023scalable, chen2023mitigating, jiang2023hallucination, huang2023opera, yu2023hallucidoctor, zhao2023beyond}, there is a noticeable gap in the literature regarding the study of behavioral hallucination. Behavioral hallucination refers to scenarios where the generated content contains actions that conflict with what is depicted in image sequences. Moreover, the existing body of work does not offer a dedicated metric for evaluating behavioral hallucinations, an area that warrants further exploration and development.
\section{Conclusion and Future Works}

In this paper, we present \ourset, an novel and challenging benchmark designed to assess the reasoning abilities of Multimodal Large Language Models (MLLMs) in interpreting image sequences. 
We conduct evaluations on nine most recent MLLMs using GPT-4-assisted evaluation procedure. 
Our findings indicate that all tested MLLMs struggle with significant behavioral and object hallucinations in generating descriptions for image sequences. 
Through a mix of quantitative analysis and case studies, we identify three primary factors contributing to these reasoning failures in MLLMs.

Looking ahead, there are three potential avenues for future research:
\textit{(1) Dataset Diversification:} We suggest further enriching \ourset by including a broader variety of data types. This expansion could encompass first-person navigation experiences, sequential medical CT scans, and interactive gaming data. Such diversification would provide a more comprehensive platform for evaluating MLLMs across a wider range of contexts and scenarios.
\textit{(2) Evaluation Process Optimization:} Another key area for development is refining the evaluation process. This involves exploring more nuanced methods to assess the reasoning capabilities of MLLMs, focusing on semantic understanding rather than relying predominantly on keyword matching. Such advancements would enable a deeper and more accurate appraisal of MLLMs' comprehension skills.
\textit{(3) Hallucination Mitigation and Reasoning Enhancement:} Lastly, informed by the three identified causes of reasoning failures, we propose developing targeted strategies to reduce both behavioral and object hallucinations in the future work. These methods would aim to bolster the reasoning faculties of MLLMs, making them more adept at accurately interpreting and describing complex image sequences.

% \section*{Limitations}

\section*{Acknowledgement}
Wang, Zhou, Liu, Xu, and Huang are supported by National Science Foundation NSF-IIS FAI program, DOD-ONR-Office of Naval Research under award number N00014-22-1-2335, DOD-AFOSR-Air Force Office of Scientific Research under award number FA9550-23-1-0048, Capital One and JP Morgan faculty fellowships. Yao thanks Center for AI Safety and Google Cloud Research Credits program for supporting our computing needs. Bansal is supported by DARPA ECOLE Program No. HR00112390060 and ONR Grant N00014-23-1-2356.

\bibliography{custom}

\appendix
\section{Details of Open X-Embodiment Data Selection}
\label{app_rtx}

In this section, we provide the names of all subsets selected from Open X-Embodiment dataset and the corresponding sampling video numbers. For detailed information, please refer to Table~\ref{tab: num_rtx}.

\begin{table*}[!htb]
\setlength{\abovecaptionskip}{0.2cm}
\centering
\caption{Number of videos selected from each sub-dataset of Open X-Embodiment.}
\begin{tabular}{l|c}
\toprule 
 Sub-dataset name & Number of videos selected \\
\midrule
fractal20220817\_data & 400 \\
kuka & 50 \\
bridge & 300\\
jaco\_play & 50\\
berkeley\_autolab\_ur5 & 50\\
toto & 10\\
columbia\_cairlab\_pusht\_real & 5\\
stanford\_hydra\_dataset\_converted\_externally\_to\_rlds & 5\\
ucsd\_kitchen\_dataset\_converted\_externally\_to\_rlds & 50\\
bc\_z & 50\\
utokyo\_pr2\_opening\_fridge\_converted\_externally\_to\_rlds & 5\\
utokyo\_pr2\_tabletop\_manipulation\_converted\_externally\_to\_rlds & 10\\
utokyo\_xarm\_pick\_and\_place\_converted\_externally\_to\_rlds & 1\\
utokyo\_xarm\_bimanual\_converted\_externally\_to\_rlds & 5\\
dlr\_sara\_pour\_converted\_externally\_to\_rlds & 5\\
dlr\_edan\_shared\_control\_converted\_externally\_to\_rlds & 100\\
asu\_table\_top\_converted\_externally\_to\_rlds & 20\\
utaustin\_mutex & 30\\
berkeley\_fanuc\_manipulation & 30\\
\bottomrule
\end{tabular}
\label{tab: num_rtx}
\end{table*}

\section{Human Evaluation}
\label{app: huamneval}
In this section, to verify the reliability of the GPT-4 assisted evaluation procedure, we compare the results of GPT-4 assisted evaluation with those of human evaluation. 
We randomly select 200 image sequences from the entire Val set and manually extract object and behavior keyword lists for each image sequence's AI-generated description and human-annotated description. 
Then, we calculate six metrics and compare them with the metrics obtained using keyword lists extracted by GPT-4. 
We choose the four MLLMs that performed best in reasoning on \ourset as representatives: GPT-4V (s-input), Gemini (s-input), Chat-UniVi (s-input), and LLaVA-1.5. 
The evaluation results are shown in Table~\ref{tab: huamn_eval}.

After comparison, we find that there is not a significant gap between the results of GPT-4 assisted evaluation and human evaluation, with the absolute value of the difference mostly ranging between 1\% to 4\%. 
For most metrics, the GPT-4 assisted evaluation tends to overestimate the performance of MLLMs, meaning the evaluation results are higher than those of human evaluation. 
However, the relative ranking among different MLLMs remains essentially unchanged. 
Overall, the GPT-4 assisted evaluation is quite reliable.

\begin{table*}[!htb]
\small
\setlength{\abovecaptionskip}{0.2cm}
\centering
\caption{Human evaluation.
% \fhc{Can you provide more detailed descriptions, in the table, of what are the kinds of videos in the daily-life, robotics and comics categories respectively?}
}
\begin{tabular}{l|c|ccc|ccc}
\toprule 
\multirow{2}{*}{\textbf{Model}} & \multirow{2}{*}{\textbf{Eval type}} &       & \textbf{Object}     &          &       & \textbf{Behavior} &  \\ 
            &               & Recall & Precision & F1 & Recall & Precision & F1\\
\midrule
\multirow{2}{*}{GPT-4V (s-input)} & GPT-4 & 60.91\% & 51.04\% & 54.13\% & 38.02\% & 33.05\% & 34.12\% \\
\cmidrule{2-8}
                                  & Human & 57.69\% & 49.54\% & 52.01\% & 35.26\% & 31.60\% & 32.67\% \\ 
\midrule
\multirow{2}{*}{Gemini (s-input)} & GPT-4 & 37.54\% & 39.43\% & 36.88\% & 23.38\% & 34.19\% & 24.02\% \\
\cmidrule{2-8}
                                  & Human & 35.82\% & 38.11\% & 37.09\% & 20.46\% & 33.72\% & 22.99\% \\
\midrule
\multirow{2}{*}{ChatUnivi (s-input)} & GPT-4 & 40.32\% & 42.04\% & 39.52\% & 24.95\% & 28.06\% & 27.15\% \\
\cmidrule{2-8}
                                  & Human & 37.65\% & 38.59\% & 36.46\% & 25.73\% & 27.40\% & 26.64\% \\
\midrule
\multirow{2}{*}{LLaVA-1.5 (c-input)} & GPT-4 & 35.77\% & 44.18\% & 38.09\% & 24.47\% & 38.79\% & 28.59\% \\
\cmidrule{2-8}
                                  & Human & 36.84\% & 41.37\% & 39.77\% & 22.95\% & 39.82\% & 29.18\% \\
\bottomrule
\end{tabular}
\label{tab: huamn_eval}
\end{table*}
\section{Prompt Details}
\label{app_prompt}
In this section, we provide all the prompts used in our paper, including those used to merge questions and answers from Daily-life videos into a single description, prompts for MLLMs to generate descriptions corresponding to image sequences, and prompts for extracting object and behavior keywords from both human-annotated and AI-generated descriptions. The detailed prompts are showm in Table~\ref{tab: prompt}.

\begin{table*}[!htb]
\setlength{\abovecaptionskip}{0.2cm}
\centering
\caption{All prompts used in our paper.}
\begin{tabularx}{\textwidth}{X}
\toprule 
 Prompt \\
\midrule
\midrule
Task: Rewrite questions and answers into a single paragraph \\
\midrule
 Image: <Image sequence> \\
 Text: <Write a description for this image based on the following questions and answers in one paragraph. Please remember that some objects or actions in the following questions and answers may not be included in the images. Please do not include the excluded items in your description. 
 Here are the questions and answers: Question: \{\textbf{Question 1}\} Answer: \{\textbf{Answer 1}\} Question: \{\textbf{Question 2}\} Answer: \{\textbf{Answer 2}\} ... Question: \{\textbf{Question n}\} Answer: \{\textbf{Answer n}\}> \\
\midrule
\midrule
Task: Generate description for the given image sequence \\
\midrule
 Image: <Image sequence> \\
 Text: <Write a description for the given image sequence in a single paragraph, what is happening in this episode?> \\
\midrule
\midrule
Task: Extract object and behavior keywords \\
\midrule
 Text: <I will provide you two paragraphs. The first paragraph is human-composed and the second paragraph is generated by AI models. I want to evaluate the hallucination in the second paragraph. Please extract the object and action words or phrases from the following text. The objects should have a tangible meaning and consist of no more than two words; non-tangible objects should not be extracted. The action words or phrases should only relate to the extracted objects. Also, you must convert the corresponding actions to their complete root form. Then, for the final answer, please examine 4 lists and must transfer the synonyms in 4 lists into the same word. Please directly output the final object and action lists in two paragraphs, respectively as in the form in the example below without any justifications or intermediate steps. \\
 Here is an example: \\
 1. The sequence of images captures a dog's cautious interaction with a metal toy inside a house. The dog appears wary and maintains a distance from the unfamiliar object, barking to express its disapproval and possibly intimidation. As the toy moves, the dog's reaction is to bark and lean backward, showing a clear sign of being unsettled by the toy's motion. When the toy momentarily ceases movement, the dog also stops, remaining alert and attentive. At the end of the image, when the toy comes to a halt, the dog looks up, still processing the strange encounter with the inanimate object. \\
 2. The image is a collage of multiple pictures featuring two dogs playing with a toy alligator. The dogs are in various positions, with some of them standing on the toy alligator, while others are interacting with it in different ways. The collage captures the dogs' playfulness and excitement as they engage with the toy alligator. \\
 The lists are \\
 Object list 1: [dog, toy, house] \\
 Action list 1: [interaction, bark, express intimidation, move, lean backward, stop, look up] \\
 Object list 2: [dog, toy] \\
 Action list 2: [play, stand, interaction] \\
 Here is the paragraphs: \\
 1. \{\textbf{Human-annotated description}\} \\
 2. \{\textbf{AI-generated description}\} \\
 The lists are:> \\
\bottomrule
\end{tabularx}
\label{tab: prompt}
\end{table*}
\newpage
\section{Case Study}
\label{app_case}
\begin{figure*}[!h]
\centering
\includegraphics[width=0.95\textwidth]{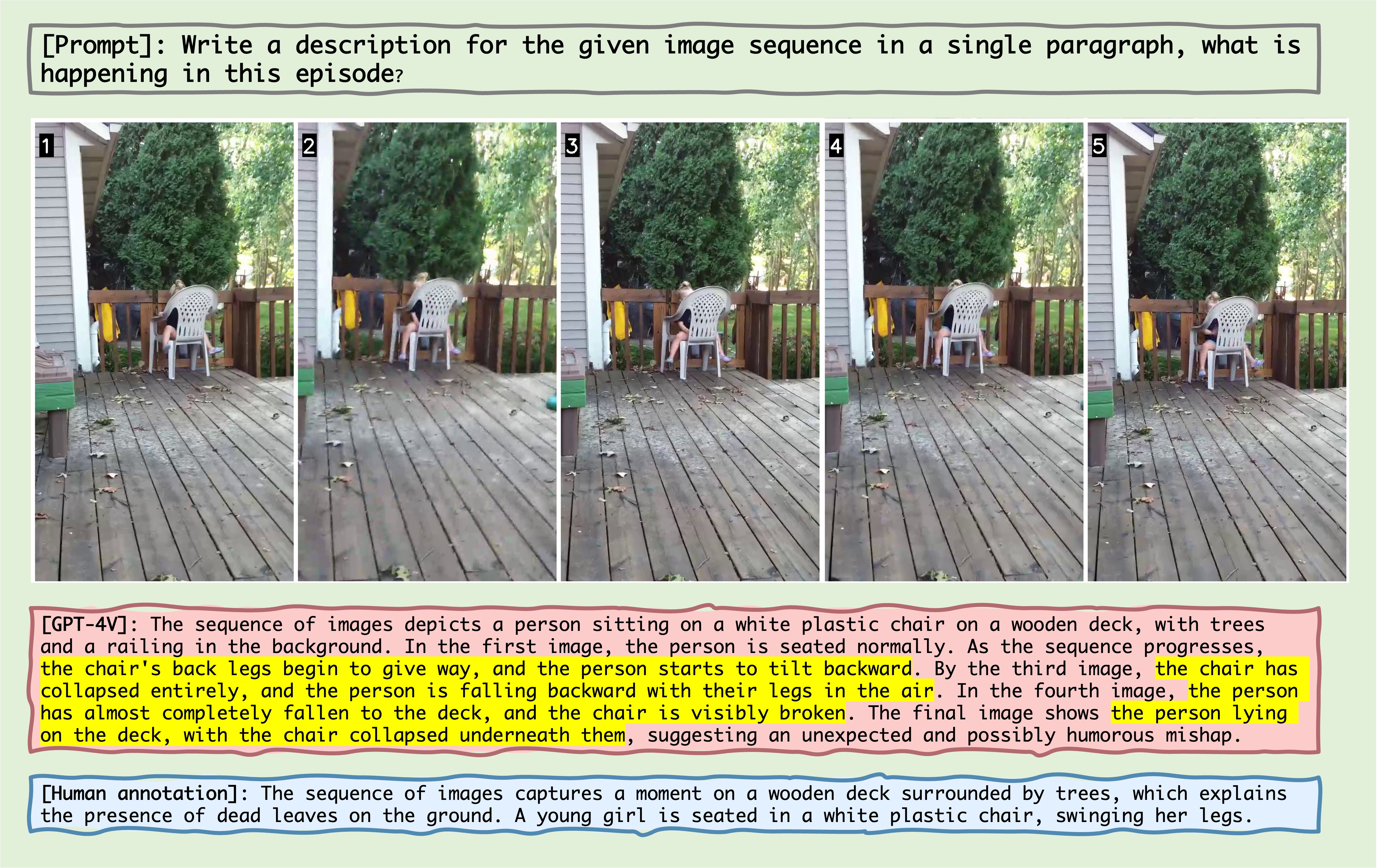}
% \vspace{-1em}
\caption{A sample of failure reasoning case in Daily-life domain, we highlight the hallucination parts in yellow. Failure reason: co-occurrence behavior and Snowball.}
\label{fig:case1}
\end{figure*}

\begin{figure*}[!h]
\centering
\includegraphics[width=0.95\textwidth]{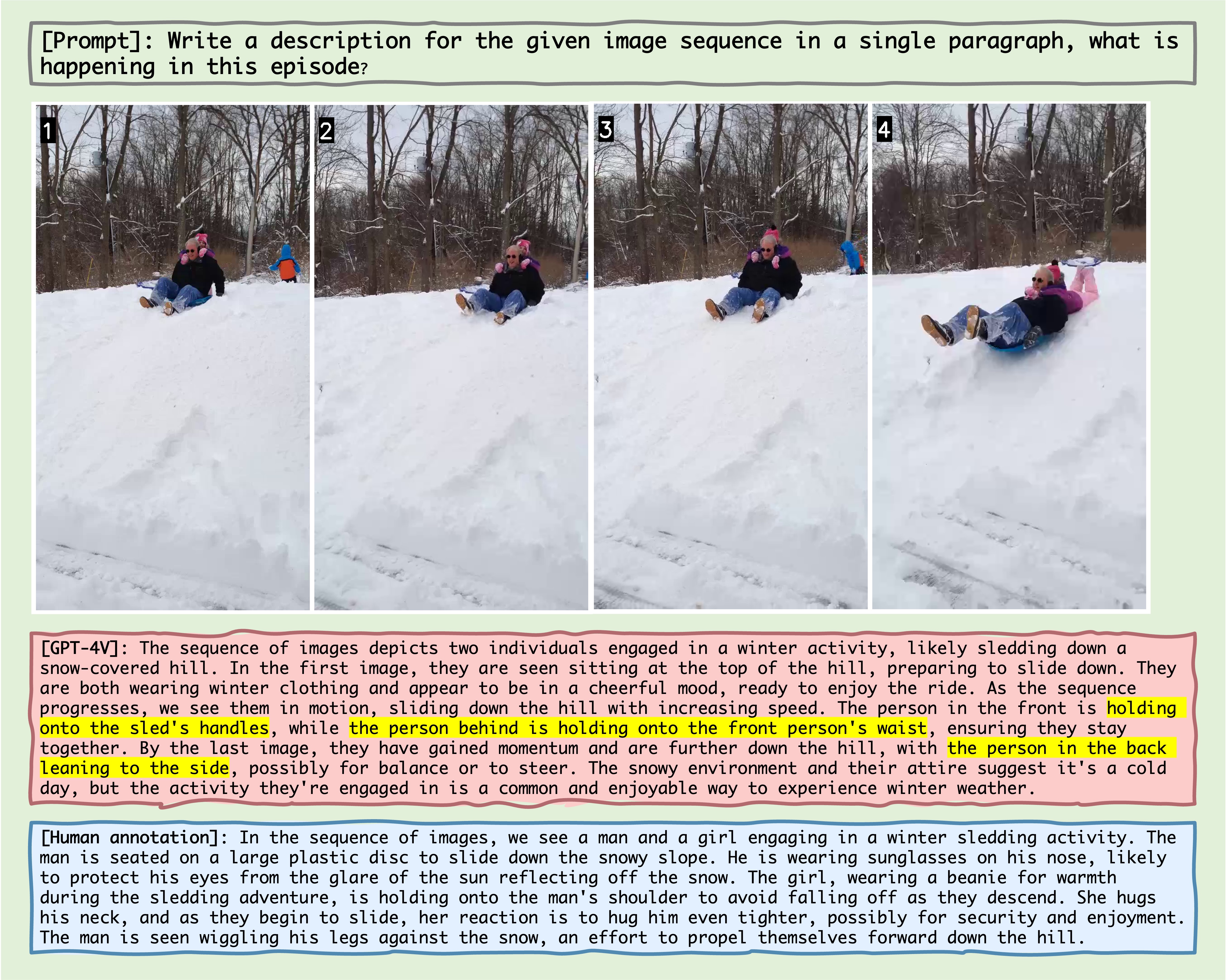}
% \vspace{-1em}
\caption{A sample of failure reasoning case in Daily-life domain, we highlight the hallucination parts in yellow. Failure reason: object hallucination, and correlation between object hallucination and behavioral hallucination.}
\label{fig:case2}
\end{figure*}

\begin{figure*}[!h]
\centering
\includegraphics[width=0.95\textwidth]{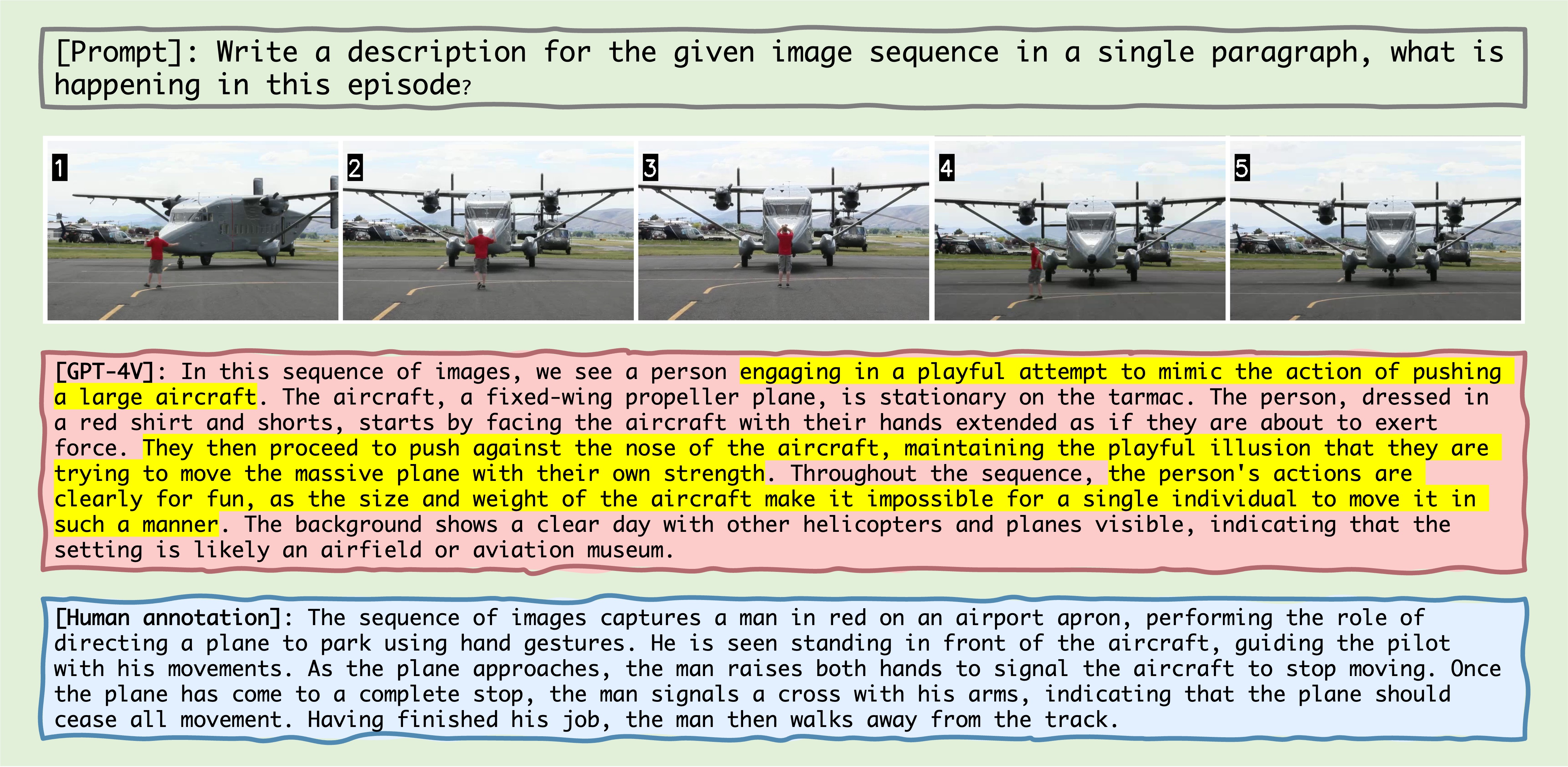}
% \vspace{-1em}
\caption{A sample of failure reasoning case in Daily-life domain, we highlight the hallucination parts in yellow. Failure reason: lack of common sense and Snowball.}
\label{fig:case3}
\end{figure*}

\begin{figure*}[!h]
\centering
\includegraphics[width=0.95\textwidth]{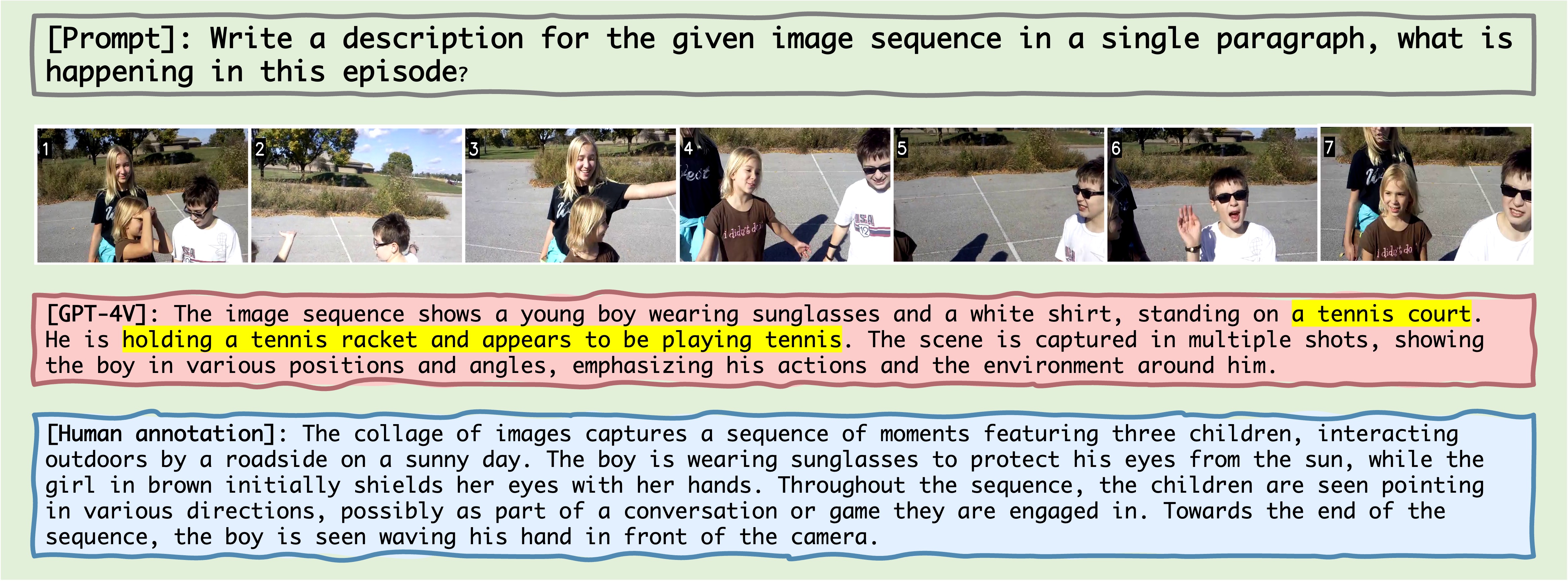}
% \vspace{-1em}
\caption{A sample of failure reasoning case in Daily-life domain, we highlight the hallucination parts in yellow. Failure reason: object hallucination, correlation between object hallucination and behavioral hallucination, and co-occurrence behavior.}
\label{fig:case4}
\end{figure*}

\begin{figure*}[!h]
\centering
\includegraphics[width=0.95\textwidth]{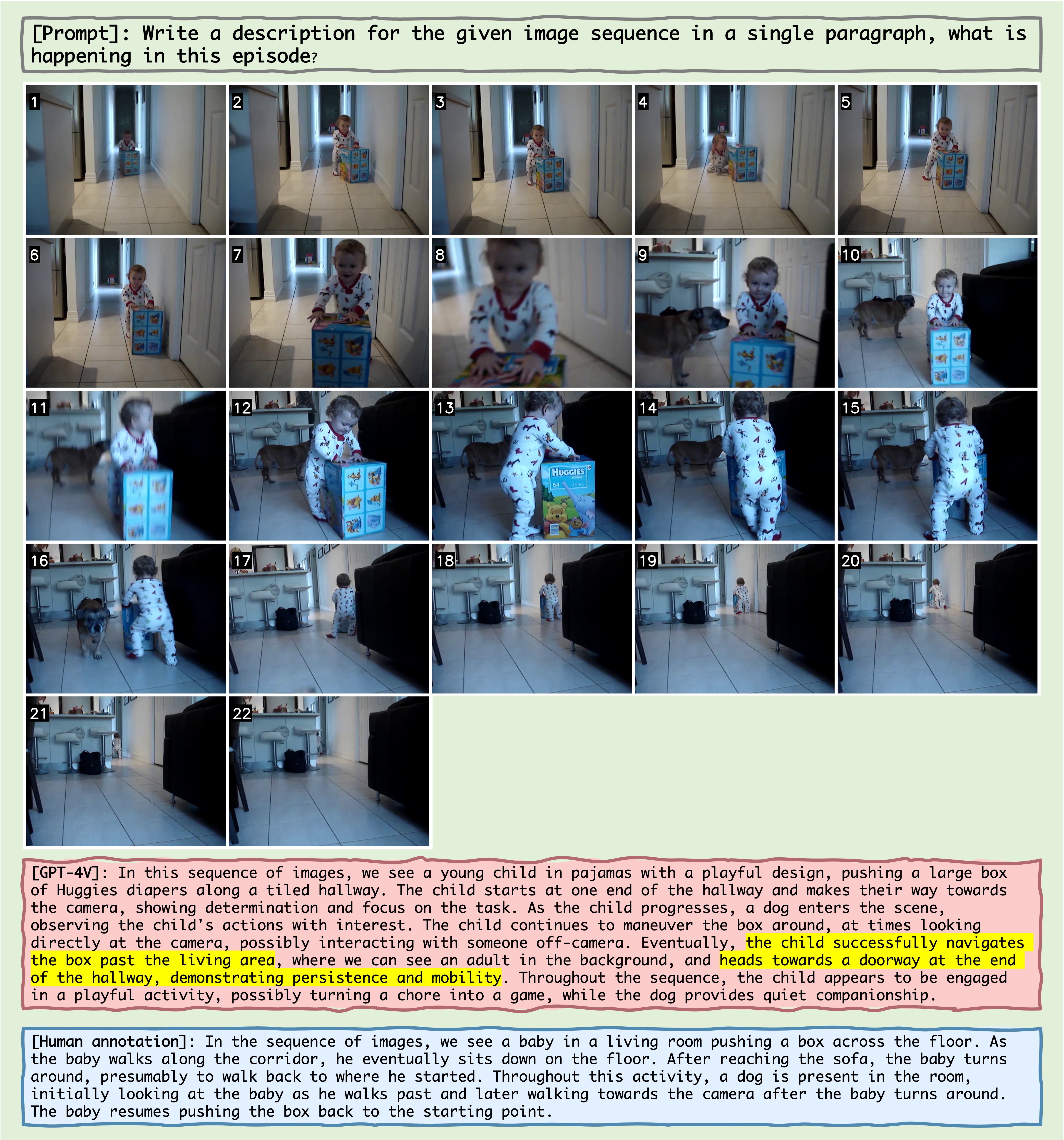}
% \vspace{-1em}
\caption{A sample of failure reasoning case in Daily-life domain, we highlight the hallucination parts in yellow. Failure reason: Snowball. In this case, we observe that in addition to the significant behavioral hallucinations caused by Snowball effect mentioned in Section~\ref{sec: error_ana}, another result of Snowball is that LVLMs may not fully describe all episodes in an image sequence. That is, after a behavioral hallucination occurs, the LVLM might assume the episode has ended and stop describing. For instance, in this case, the LVLM stopped describing after mentioning the child reaching the living room and the adult leaving, without continuing to describe the child pushing the box back along the hallway.}
\label{fig:case5}
\end{figure*}

\begin{figure*}[!h]
\centering
\includegraphics[width=0.95\textwidth]{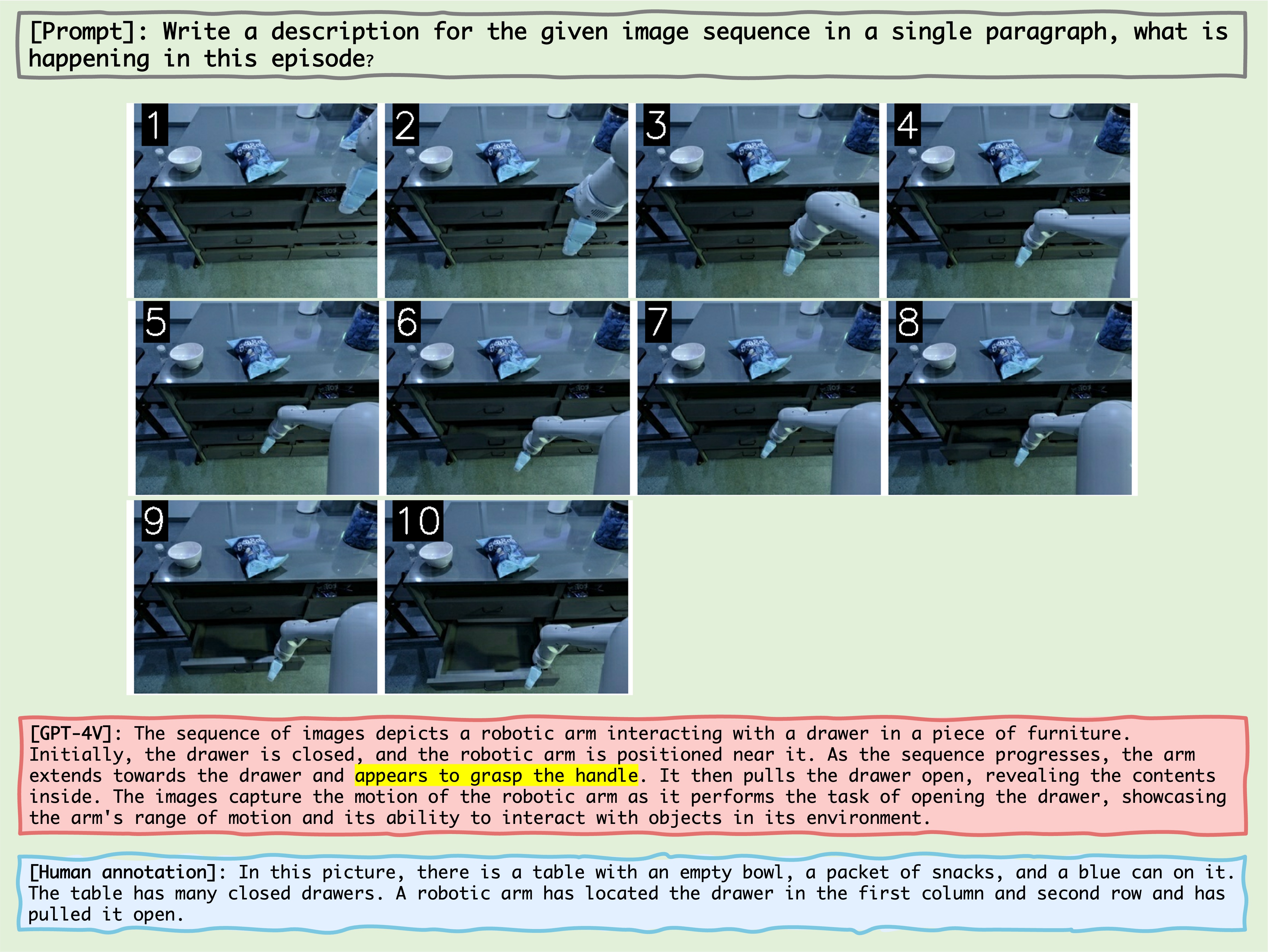}
% \vspace{-1em}
\caption{A sample of failure reasoning case in Robotics domain, we highlight the hallucination parts in yellow. Failure reason: co-occurrence behavior.}
\label{fig:case6}
\end{figure*}

\begin{figure*}[!h]
\centering
\includegraphics[width=0.95\textwidth]{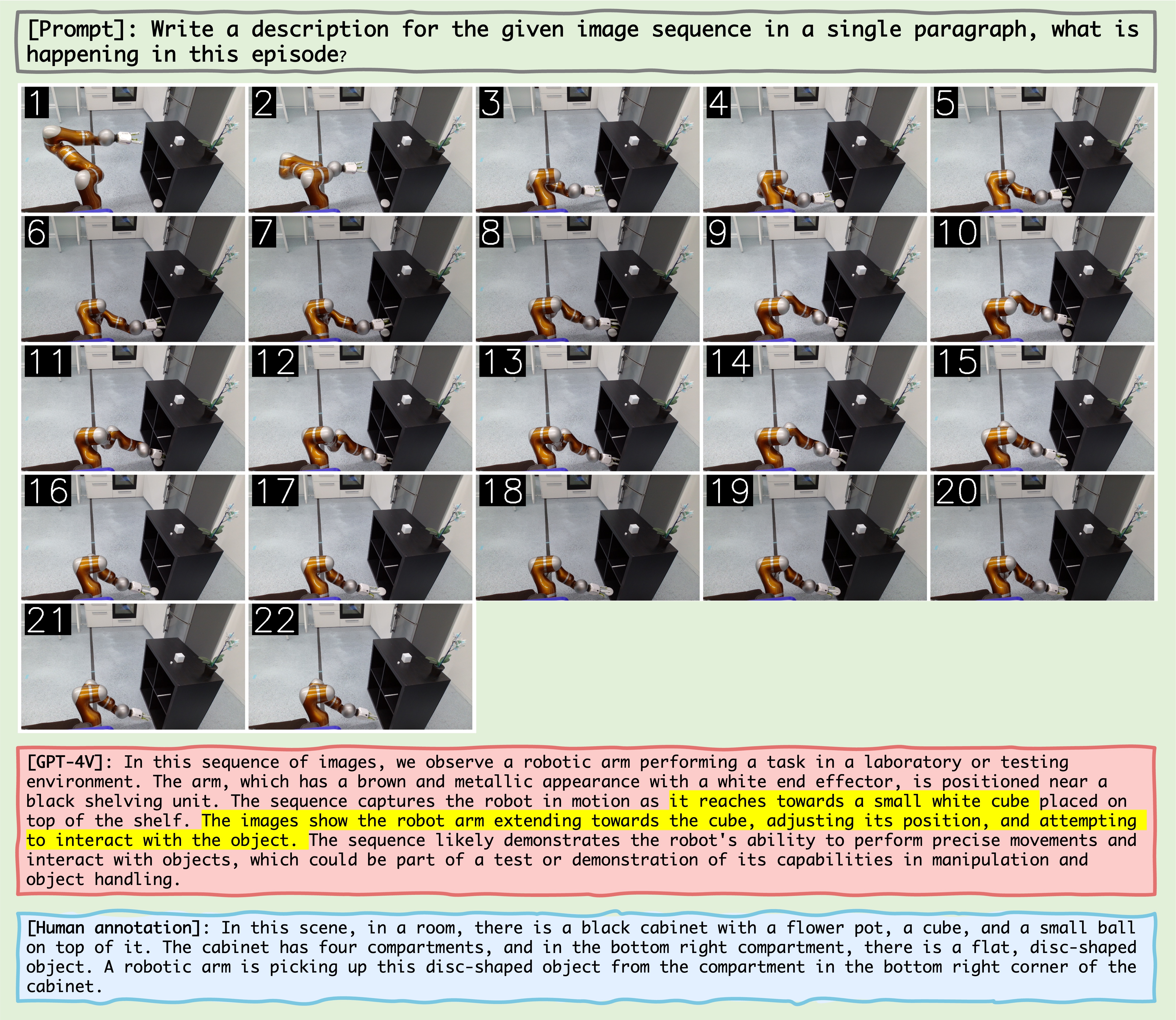}
% \vspace{-1em}
\caption{A sample of failure reasoning case in Robotics domain, we highlight the hallucination parts in yellow. Failure reason: Snowball. This case effectively demonstrates the lack of LVLM's reasoning ability in image sequence comprehension. In the first image, the robotic arm indeed appears to be moving towards the cube, but from the second image, the arm lowers and moves towards the disc-shaped object. The LVLM failed to infer this behavior from the first two images and based its subsequent description solely on the understanding in the first image, leading to a Snowball effect.}
\label{fig:case7}
\end{figure*}

\begin{figure*}[!h]
\centering
\includegraphics[width=0.95\textwidth]{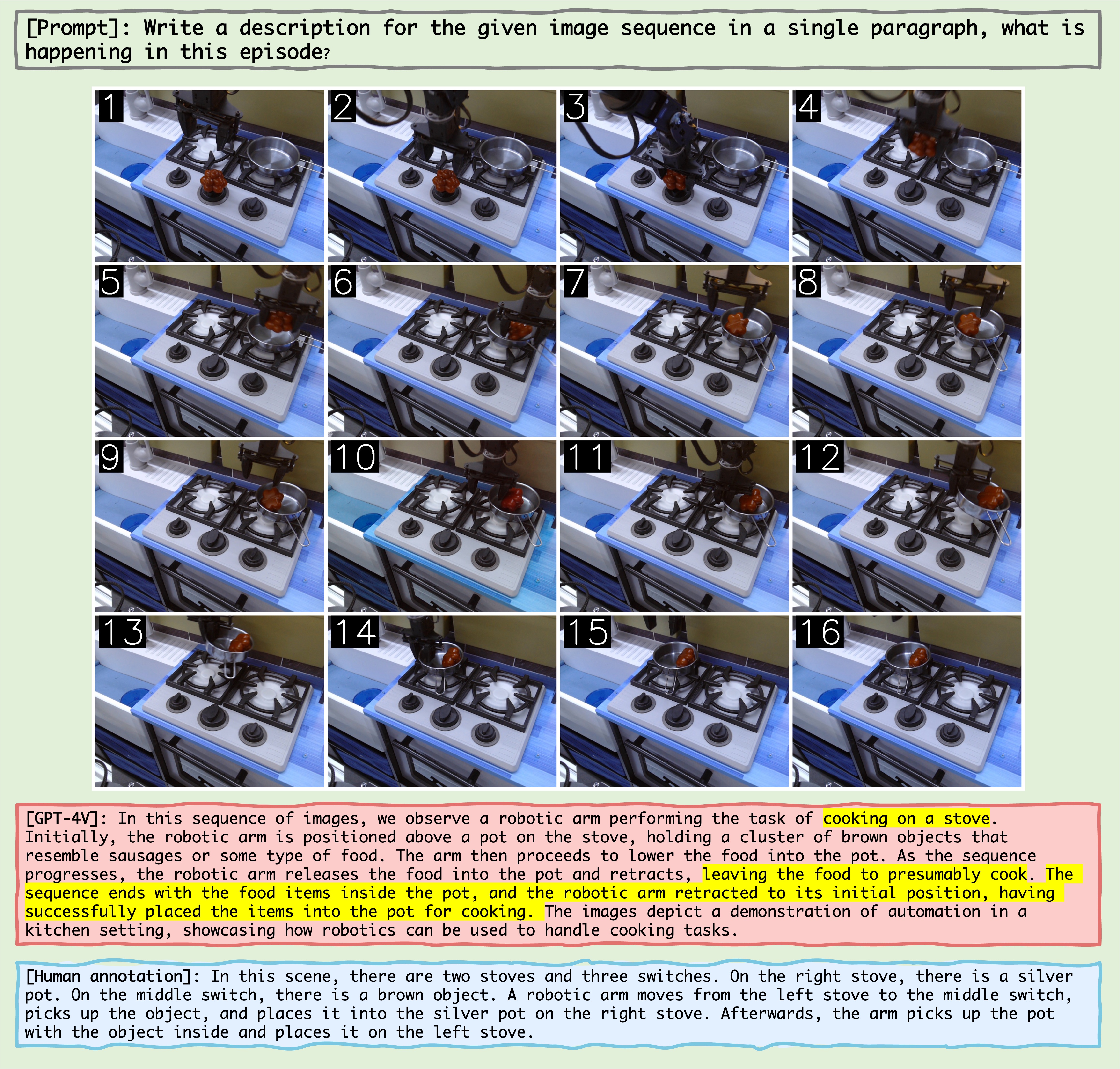}
% \vspace{-1em}
\caption{A sample of failure reasoning case in Robotics domain, we highlight the hallucination parts in yellow. Failure reason: co-occurrence behavior and Snowball. This case also reflects another outcome of the Snowball effect that we mentioned in Figure~\ref{fig:case5}. After assuming that the robotic arm is cooking, the LVLM do not continue to describe the behavior of the robotic arm moving the pot from the right stove to the left.}
\label{fig:case8}
\end{figure*}

\begin{figure*}[!h]
\centering
\includegraphics[width=0.95\textwidth]{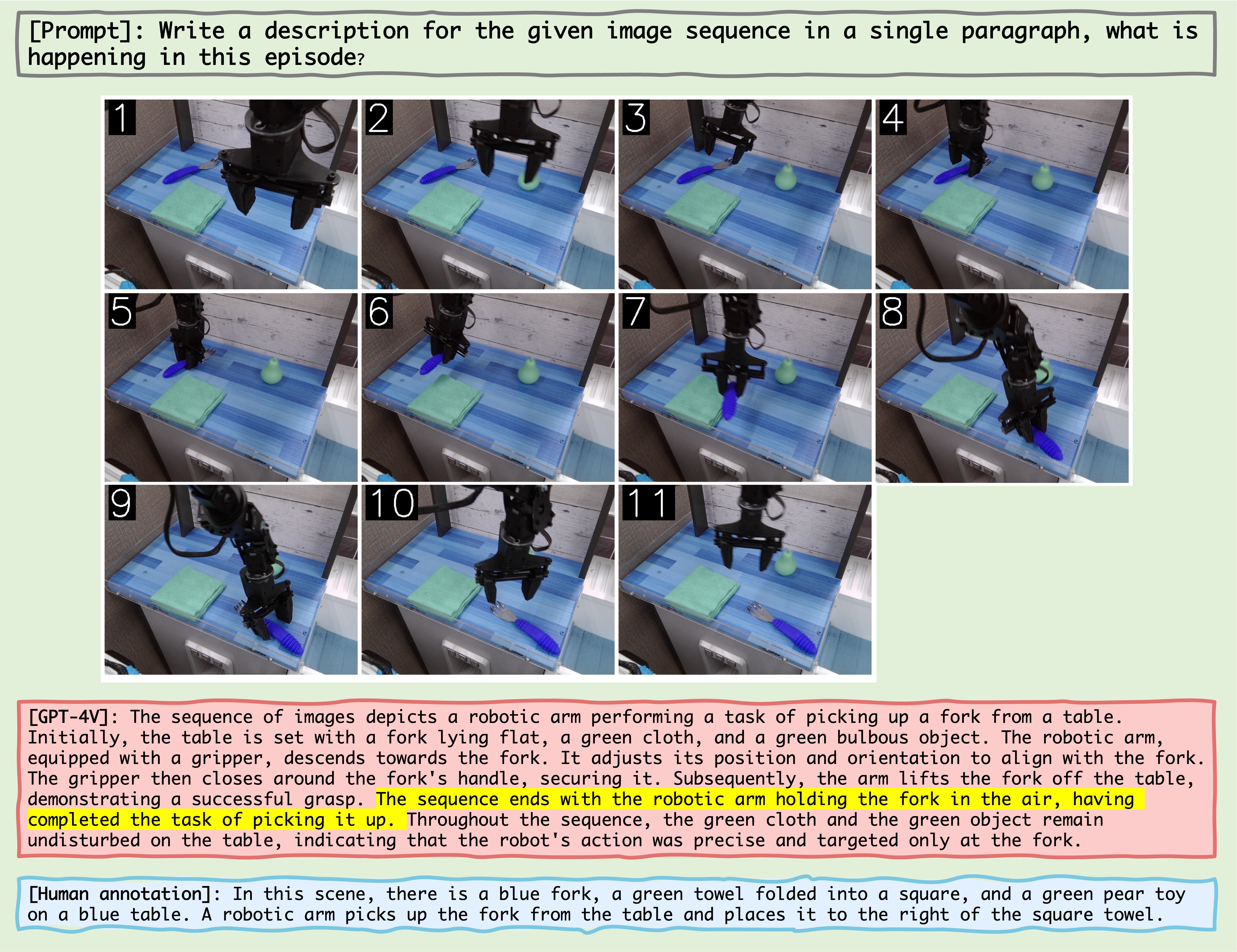}
% \vspace{-1em}
\caption{A sample of failure reasoning case in Robotics domain, we highlight the hallucination parts in yellow. Failure reason: Snowball.}
\label{fig:case9}
\end{figure*}

\begin{figure*}[!h]
\centering
\includegraphics[width=0.95\textwidth]{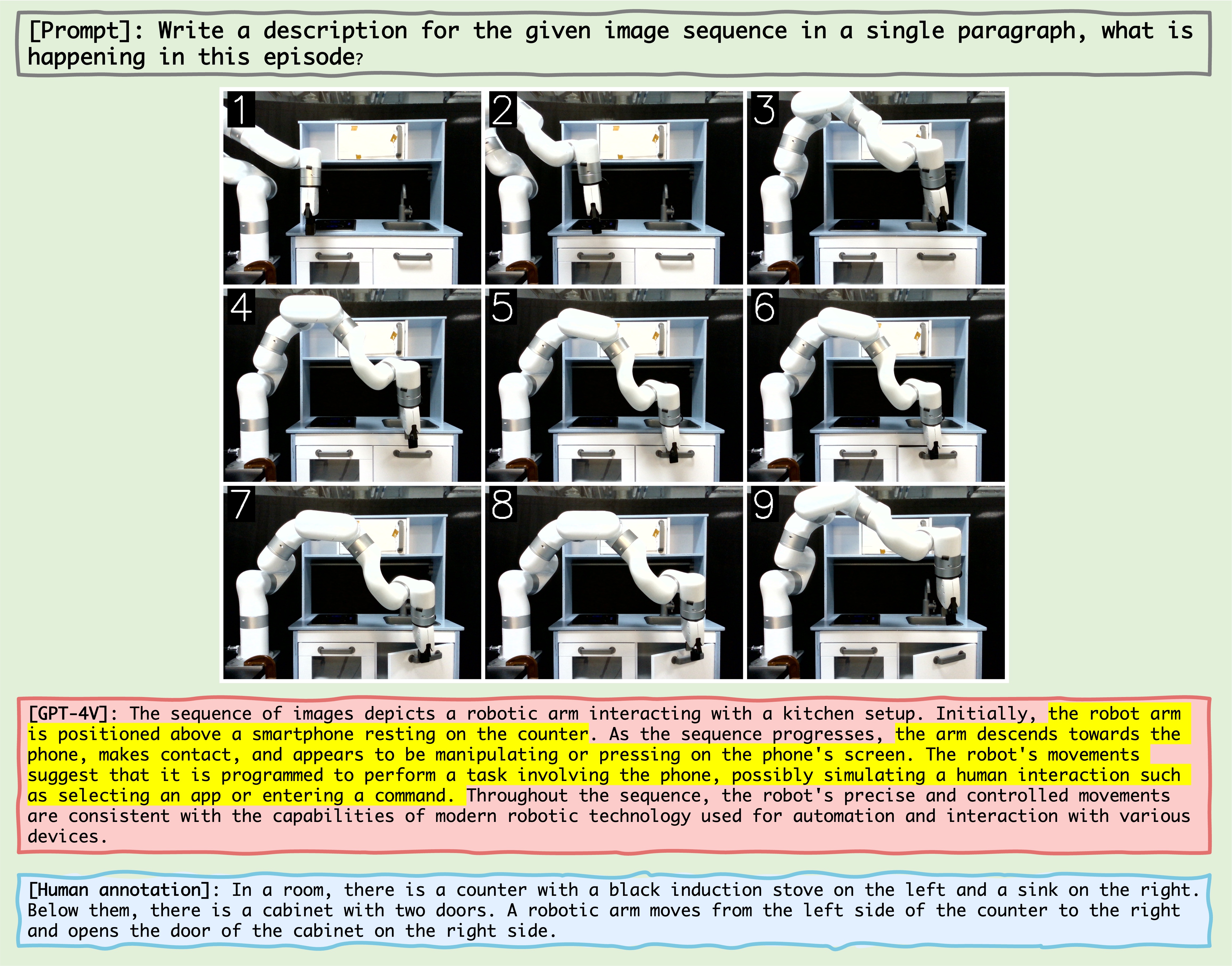}
% \vspace{-1em}
\caption{A sample of failure reasoning case in Robotics domain, we highlight the hallucination parts in yellow. Failure reason: object hallucination, correlation between object hallucination and behavioral hallucination, and Snowball.}
\label{fig:case10}
\end{figure*}

\begin{figure*}[!h]
\centering
\includegraphics[width=0.95\textwidth]{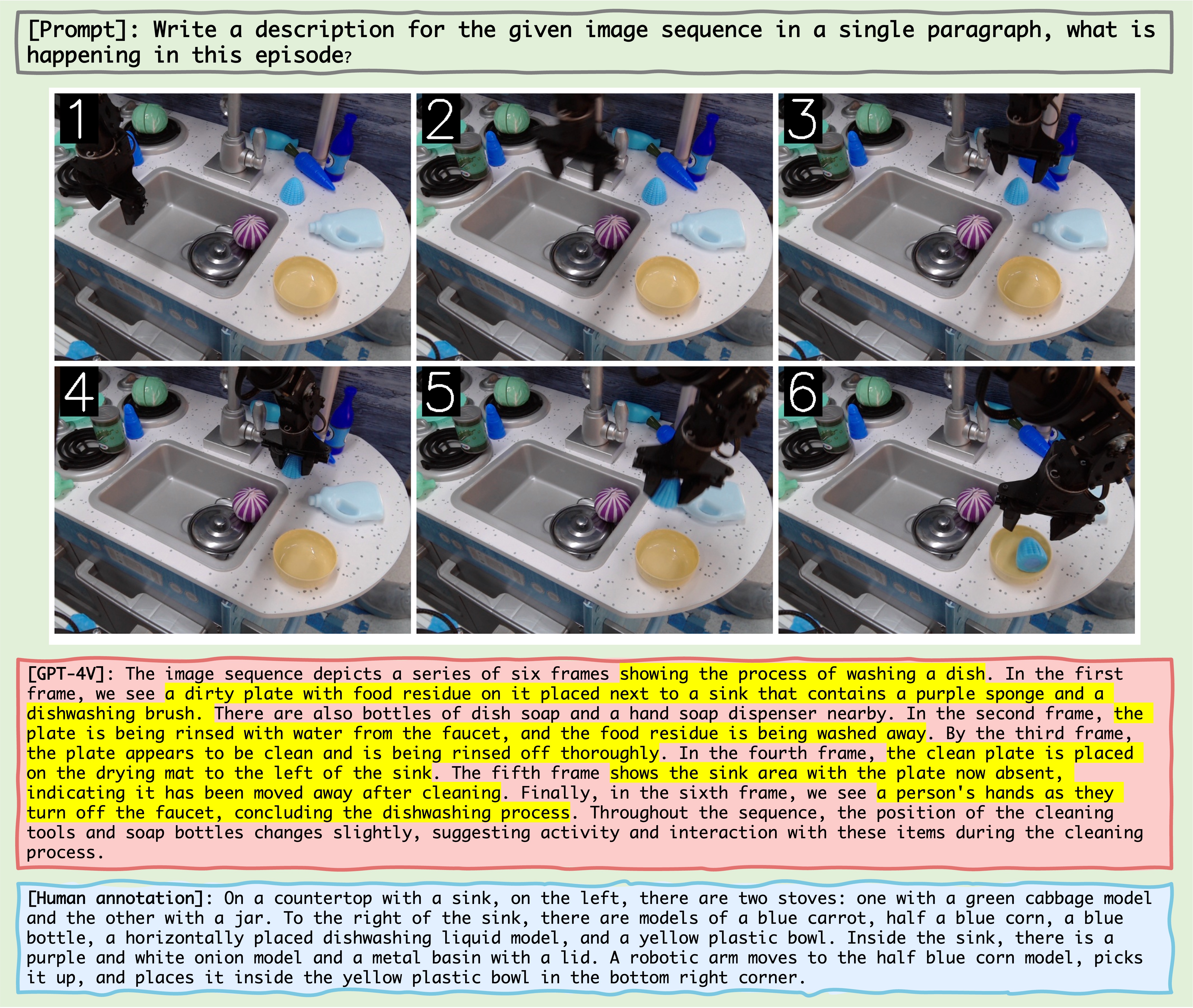}
% \vspace{-1em}
\caption{A sample of failure reasoning case in Robotics domain, we highlight the hallucination parts in yellow. Failure reason: object hallucination, correlation between object hallucination and behavioral hallucination, and Snowball.}
\label{fig:case11}
\end{figure*}

\begin{figure*}[!h]
\centering
\includegraphics[width=0.95\textwidth]{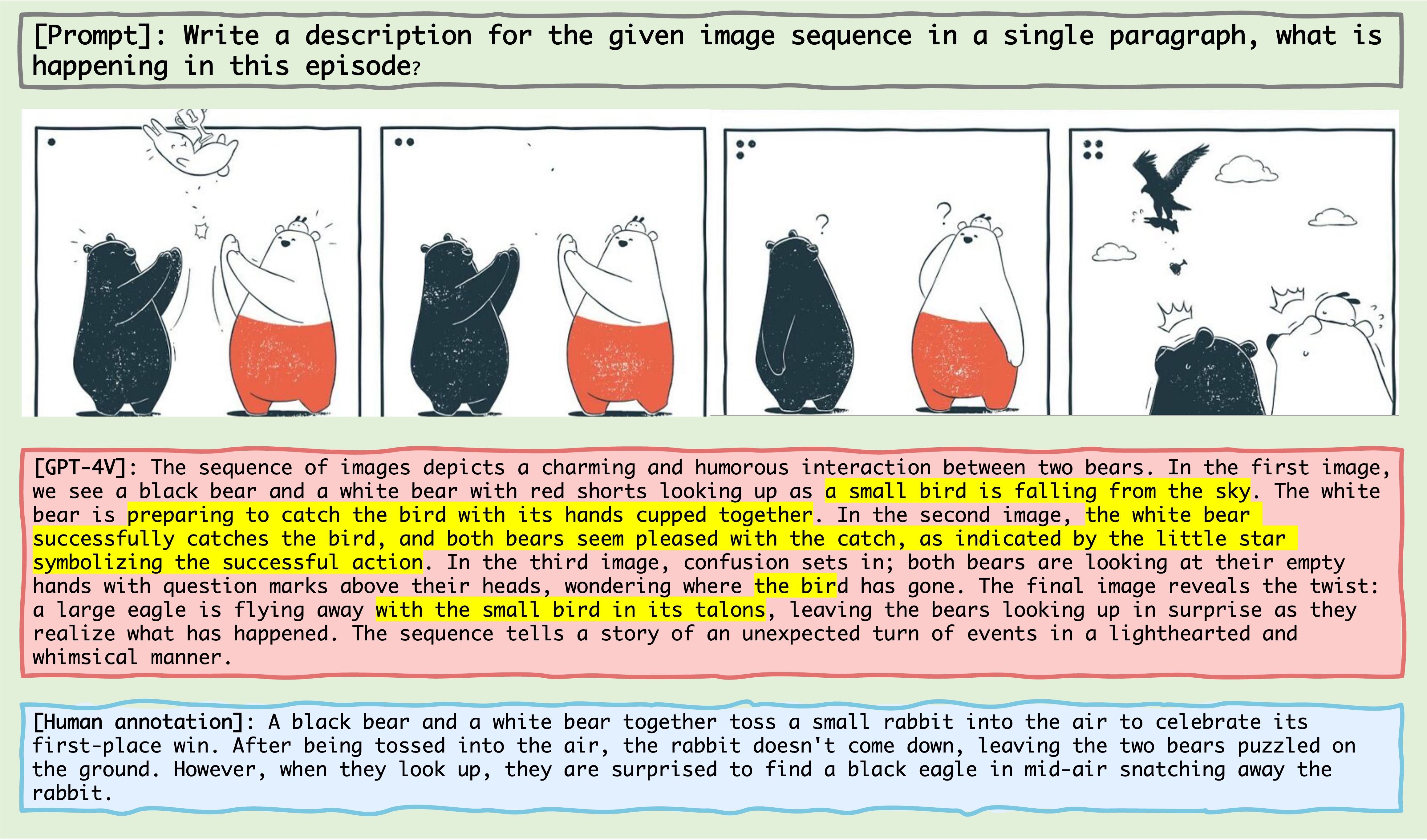}
% \vspace{-1em}
\caption{A sample of failure reasoning case in Comics domain, we highlight the hallucination parts in yellow. Failure reason: object hallucination, correlation between object hallucination and behavioral hallucination, and Snowball.}
\label{fig:case12}
\end{figure*}

\begin{figure*}[!h]
\centering
\includegraphics[width=0.95\textwidth]{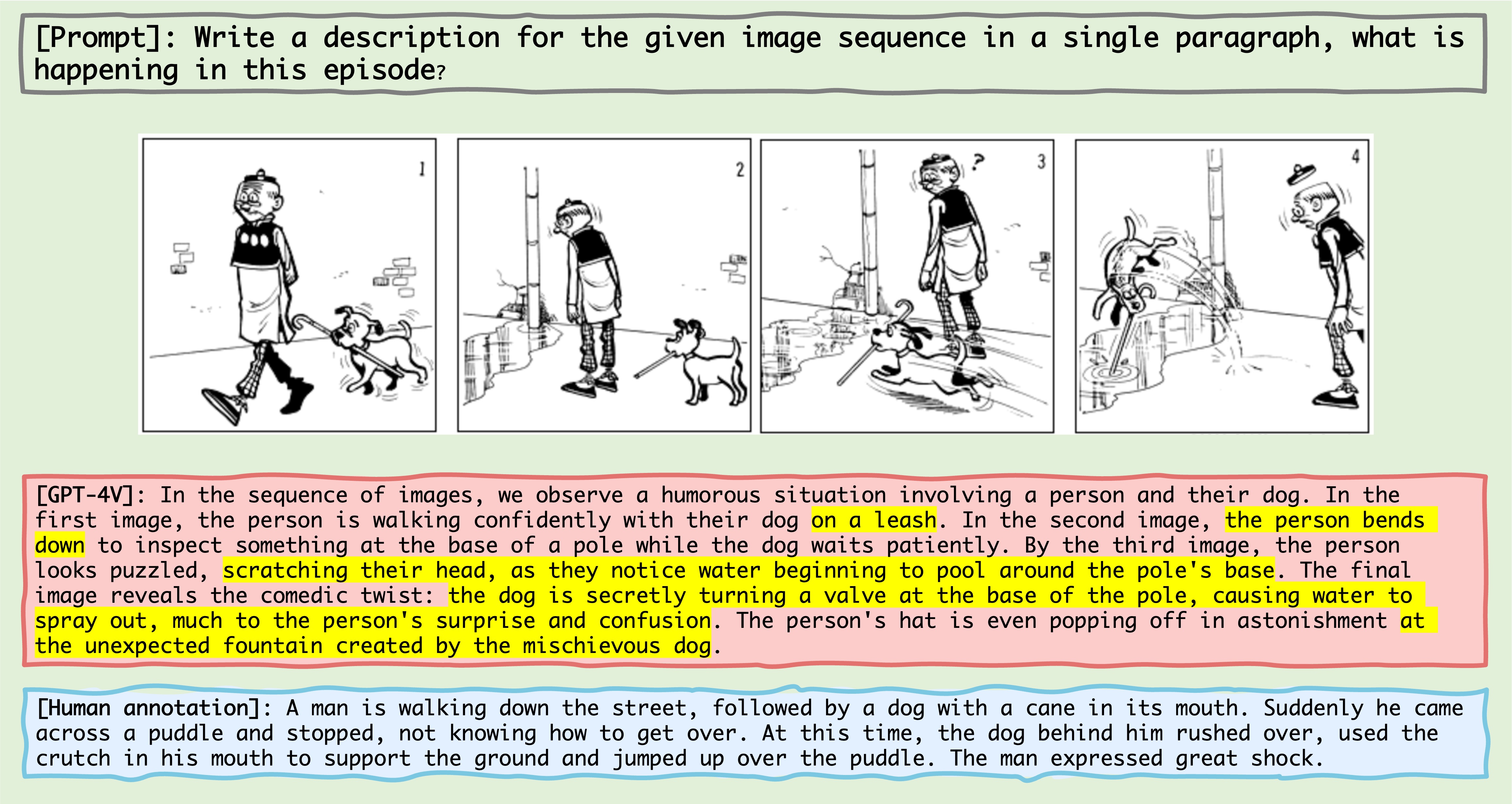}
% \vspace{-1em}
\caption{A sample of failure reasoning case in Comics domain, we highlight the hallucination parts in yellow. Failure reason: object hallucination, correlation between object hallucination and behavioral hallucination, and Snowball.}
\label{fig:case13}
\end{figure*}

\begin{figure*}[!h]
\centering
\includegraphics[width=0.95\textwidth]{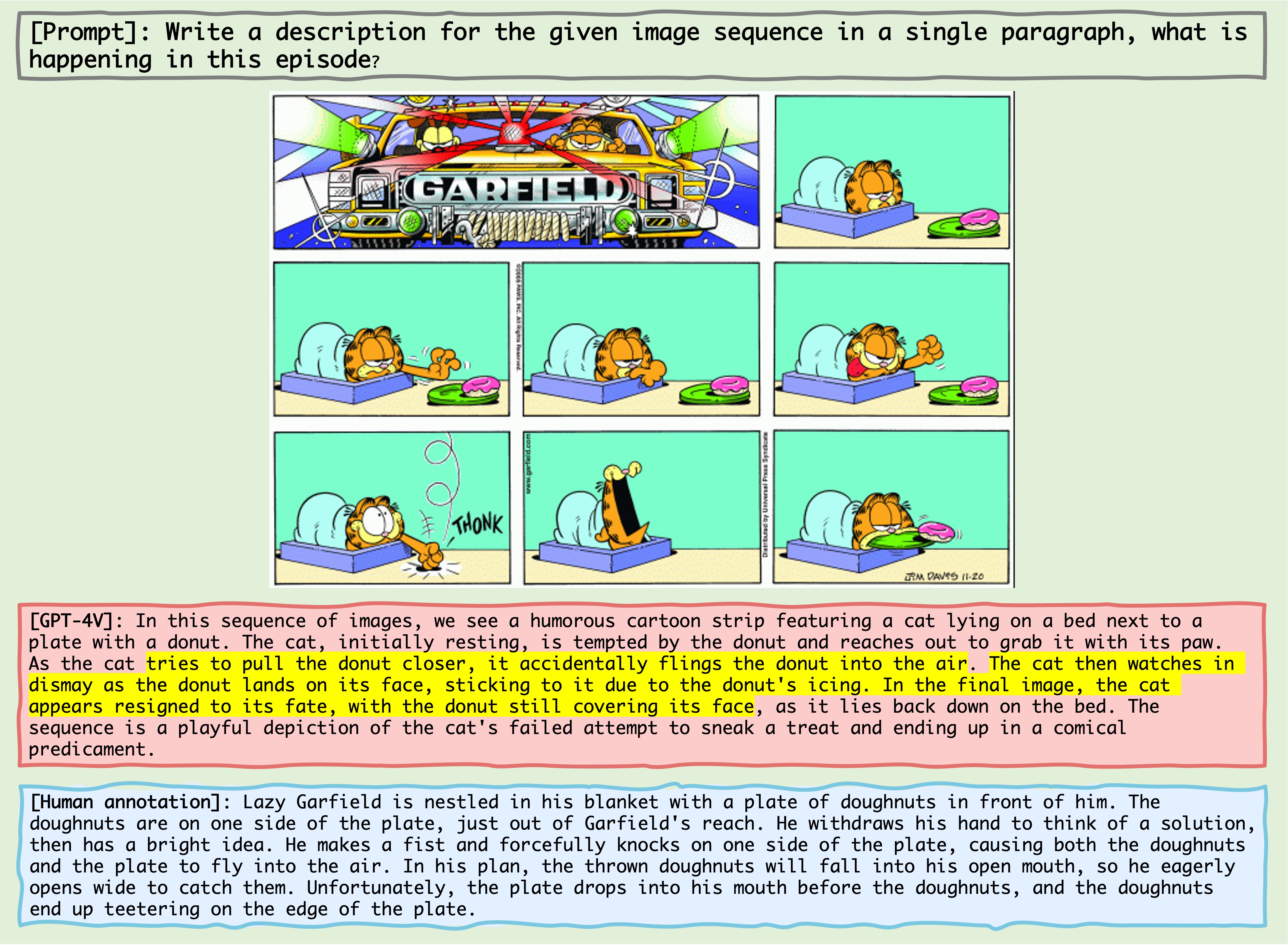}
% \vspace{-1em}
\caption{A sample of failure reasoning case in Comics domain, we highlight the hallucination parts in yellow. Failure reason: Snowball.}
\label{fig:case14}
\end{figure*}

\begin{figure*}[!h]
\centering
\includegraphics[width=0.95\textwidth]{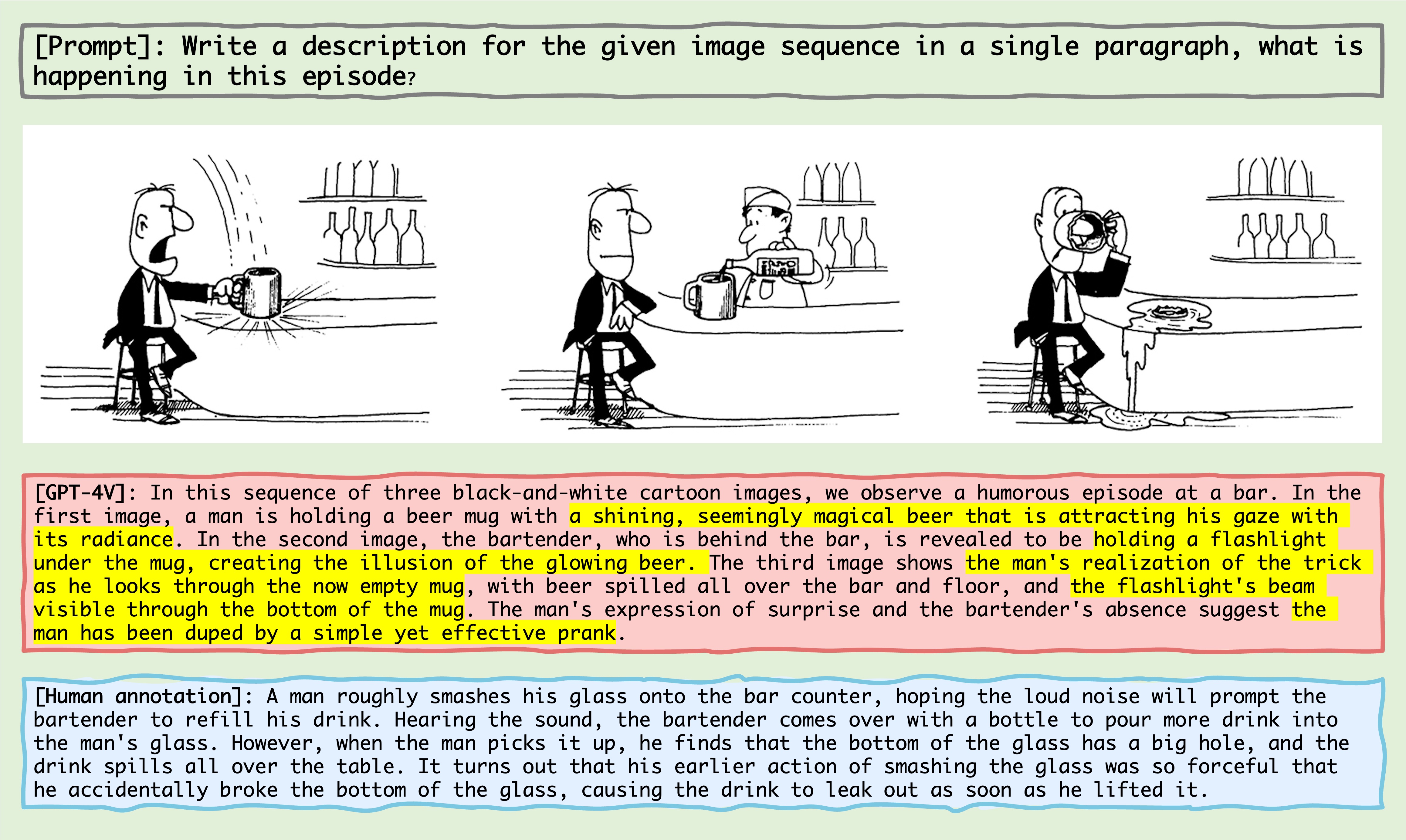}
% \vspace{-1em}
\caption{A sample of failure reasoning case in Comics domain, we highlight the hallucination parts in yellow. Failure reason: object hallucination, correlation between object hallucination and behavioral hallucination, and Snowball.}
\label{fig:case15}
\end{figure*}

\begin{figure*}[!h]
\centering
\includegraphics[width=0.95\textwidth]{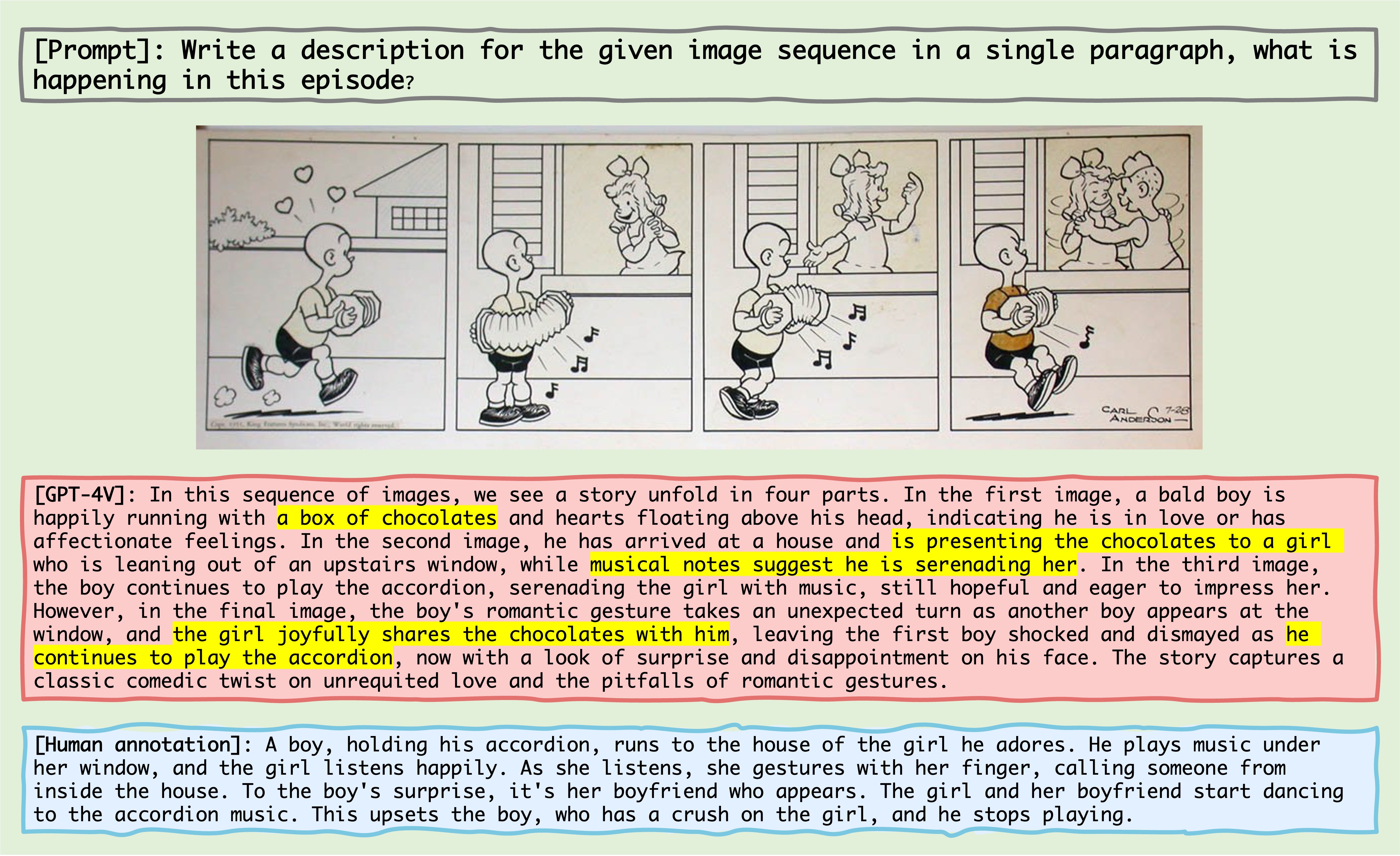}
% \vspace{-1em}
\caption{A sample of failure reasoning case in Comics domain, we highlight the hallucination parts in yellow. Failure reason: object hallucination, correlation between object hallucination and behavioral hallucination, and Snowball.}
\label{fig:case16}
\end{figure*}

In this section, we present failure reasoning cases of different domains (Figure~\ref{fig:case1}-\ref{fig:case16}), with specific reasons for failure detailed in the captions of each figure.

\end{document}